\documentclass[10pt,twocolumn,letterpaper]{article}

\usepackage{3dv}
\usepackage{times}
\usepackage{epsfig}
\usepackage{graphicx}
\usepackage{amsmath}
\usepackage{amssymb}

\usepackage{amsmath,amssymb} 
\usepackage{color}
\usepackage{graphicx}
\usepackage{amsmath}
\usepackage{amssymb}
\usepackage{booktabs}
\usepackage{subcaption}
\usepackage{comment}
\usepackage{dsfont}
\usepackage{xcolor}
\usepackage{enumitem}

\usepackage[pagebackref=true,breaklinks=true,colorlinks,bookmarks=false]{hyperref}
\usepackage[ruled,vlined,linesnumbered]{algorithm2e}
\newcommand{\boldv}[1]{\textbf{\textit{#1}}}

\usepackage[capitalize]{cleveref}
\crefname{section}{Sec.}{Secs.}
\Crefname{section}{Section}{Sections}
\Crefname{table}{Table}{Tables}
\crefname{table}{Tab.}{Tabs.}
\SetKwInput{KwDef}{Define}
\SetKwProg{Init}{init}{}{}

\threedvfinalcopy 

\def\threedvPaperID{85} 

\ifthreedvfinal\fi
\begin{document}

\title{The Surprising Positive Knowledge Transfer in\\ Continual 3D Object Shape Reconstruction}

\author{Anh Thai, Stefan Stojanov, Zixuan Huang, Isaac Rehg, James M. Rehg \\
Georgia Institute of Technology\\
\{athai6, sstojanov, zixuanh, isaacrehg, rehg\}@gatech.edu
}

\maketitle
\thispagestyle{empty}

\begin{abstract}
Continual learning has been extensively studied for classification tasks with methods developed to primarily avoid catastrophic forgetting, a phenomenon where earlier learned concepts are forgotten at the expense of more recent samples. In this work, we present a set of continual 3D object shape reconstruction tasks, including complete 3D shape reconstruction from different input modalities, as well as visible surface (2.5D) reconstruction which, surprisingly demonstrate positive knowledge (backward and forward) transfer when training with solely standard SGD and without additional heuristics. We provide evidence that continuously updated representation learning of single-view 3D shape reconstruction improves the performance on learned and novel categories over time. We provide a novel analysis of knowledge transfer ability by looking at the output distribution shift across sequential learning tasks. Finally, we show that the robustness of these tasks leads to the potential of having a proxy representation learning task for continual classification. The codebase, dataset and pre-trained models released with this article can be found at https://github.com/rehg-lab/CLRec
\end{abstract}

\begin{figure}[t!]
\begin{minipage}[t!]{\linewidth}
\centering
    \includegraphics[width=0.9\linewidth]{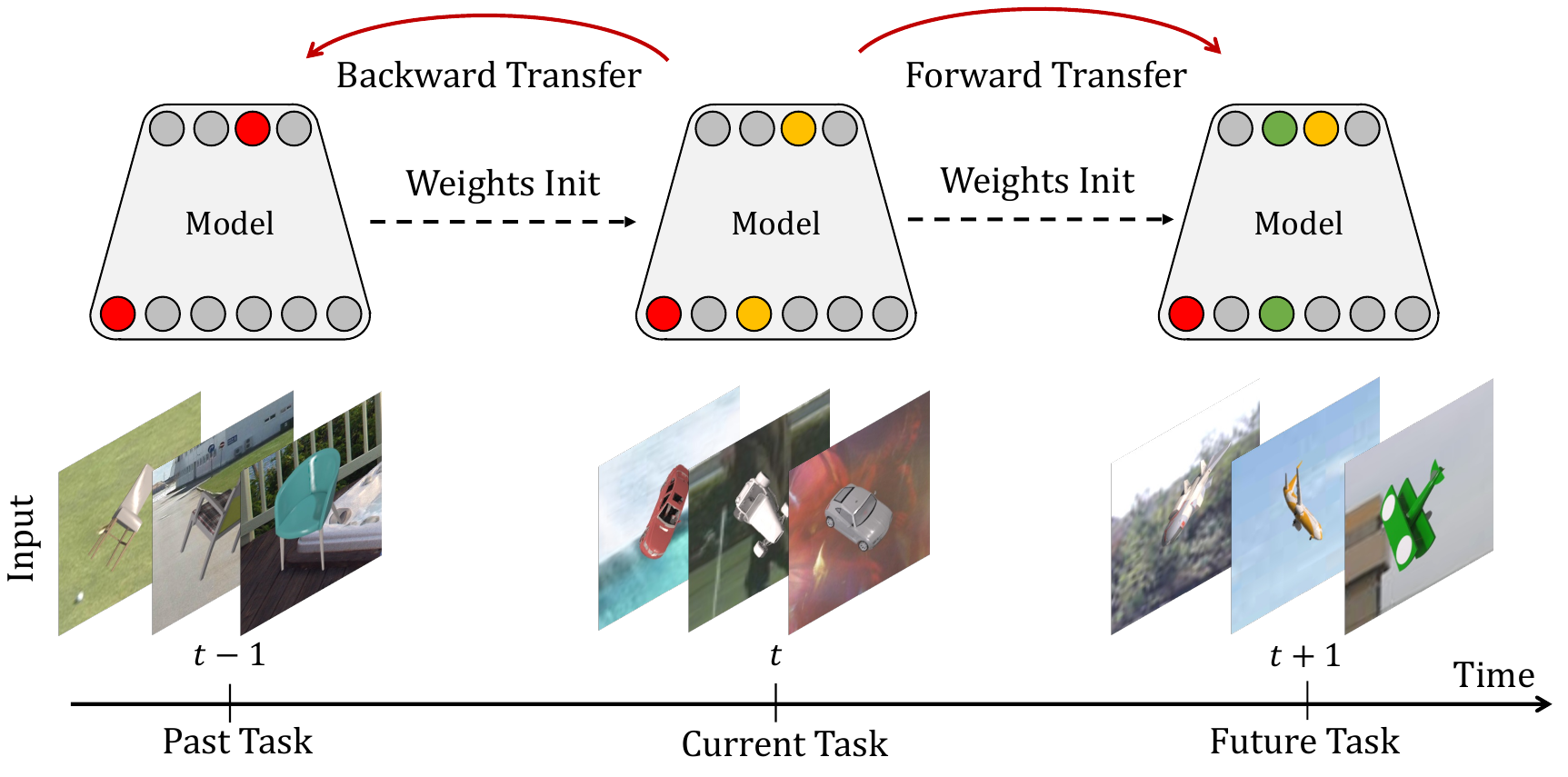}
    \vspace{-10pt}
    \caption{Continual learning setting: The model is trained continually upon receiving input data in sequence without access to past data. Backward transfer and forward transfer refer to the effect that the representations trained on the current task (yellow circles) have on solving the past tasks (red circles) and future tasks (green circles) respectively. An ideal CL learner achieves both positive backward and forward transfer.}
    \label{fig:cl}
\end{minipage}%
\vspace{-20pt}
\end{figure}
\section{Introduction}

Various applications in domains like AR/VR, autonomous driving, and robotics where 3D reconstruction is essential require continually learning and processing streams of input data. For instance, a home robot assistant learns about the newly installed bathtub after being familiar with known household objects like chairs and tables. While many important properties of 3D object shape reconstruction methods such as generalization ability and large-scale batch training have been studied extensively in prior works~\cite{tatarchenko2019single,mescheder2019occupancy,thai20203d,xu2019disn}, the feasibility of this task in a continual learning setting has not been investigated.

The goal of continual learning (CL) is to train models \emph{incrementally} to solve a sequence of tasks \emph{without} access to past data. The learner receives a sequence of learning exposures,\footnote{We use the term \emph{learning exposure} to refer to each new increment of data, i.e. the learner's next "exposure" to the concepts being learned.} each containing a subset of the overall data distribution and comprising a task (e.g., in image classification a learning exposure might contain samples from two ImageNet classes.) Note that this setting is in stark contrast to the batch training setting where the model is optimized upon observing the entire training data distribution. The fundamental challenge of CL is backward and forward knowledge transfer~\cite{lopez2017gradient}. Backward transfer (BWT) refers to the effectiveness of the current representation in solving previously-learned tasks. Large negative BWT results in catastrophic forgetting, the phenomenon where the representations learned in previous learning exposures degrade significantly over time at the expense of more recent data. For example, learning classification on 10 tasks with 20 classes/task sequentially on Tiny-ImageNet~\cite{russakovsky2015imagenet}  with solely vanilla SGD training leads to only 7.92\% average accuracy at the end, when tested on all classes. On the contrary, batch training obtains 60\%~\cite{buzzega2020dark}. Tackling catastrophic forgetting has been attempted by a large number of prior works~\cite{de2019continual,zhang2020class,liu2020mnemonics,Michieli_2019_ICCV} by employing multiple complex training heuristics and has come to characterize continual learning for many different tasks (e.g., classification, segmentation, detection, etc.) Also important is forward transfer (FWT), which refers to the utility of the learned representation for unseen future tasks. Positive FWT 
enables CL methods to leverage shared representations across tasks, so that training on new tasks is more effective than training from scratch. Past works have largely focused on classification tasks~\cite{rebuffi2016icarl,castro2018end,lopez2017gradient,prabhugdumb}, with a few exceptions~\cite{Cai_2021_ICCV,Yan_2021_ICCV}. A common theme of these efforts is the difficulty of avoiding negative BTW and achieving positive FWT. Please see Fig.~\ref{fig:cl} for an illustration of the standard CL setting.

\begin{figure*}[t!]
\begin{minipage}[t!]{\linewidth}
\centering
\vspace{-15pt}
    \includegraphics[width=0.8\linewidth]{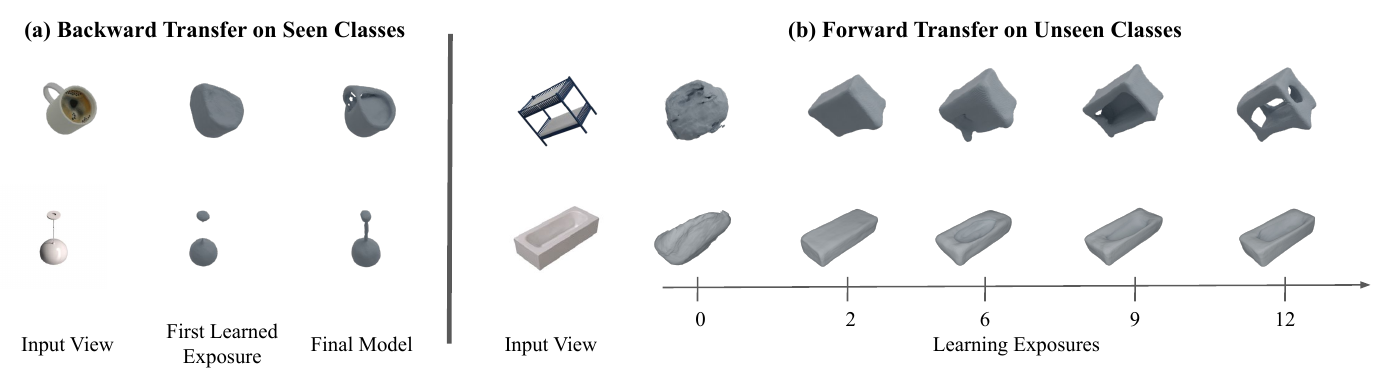}
    \vspace{-10pt}
    \caption{CL 3D shape reconstruction exhibits: (a) Positive backward transfer, reconstructions based on the first learning exposure slightly improve by the time the final model is trained, demonstrating lack of catastrophic forgetting, and (b) Positive forward transfer, reconstruction performance on unseen object classes improves steadily during training, demonstrating the generalization ability of shared representation. These results are obtained by vanilla SGD \emph{without} special architectures, losses, or exemplars. Note that each object class is only learned once in the learning exposure that it is first introduced to the model. We show more qualitative results in the Supp.}
    \label{fig:bwt_fwt}
\end{minipage}%

\vspace{-20pt}
\end{figure*}

In this work, we demonstrate that \emph{continual object shape reconstruction tasks} exhibit surprisingly effective knowledge transfer using standard deep architectures and vanilla SGD, 
\emph{without any of the special losses, exemplars, or other approaches} routinely used in CL to overcome forgetting. This is illustrated in Fig.~\ref{fig:bwt_fwt} for the challenging task of single-view 3D shape reconstruction~\cite{tatarchenko2019single,thai20203d,xu2019disn}, in which the learner predict the 3D shape of an object given a single input image. Each learning exposure contains samples from a subset of object classes,\footnote{While there is nothing inherently categorical about shape reconstruction, categories are routinely-used to identify sets of similar shapes for training and evaluation purposes, e.g. in testing generalization to unseen categories of shapes~\cite{tatarchenko2019single,thai20203d,genre}.} and we test generalization to both seen and unseen classes of objects. Fig.~\ref{fig:bwt_fwt}a illustrates the BWT performance of our CL reconstruction approach. The shape reconstructions rendered in the second column were produced after the model received its first (and only) learning exposure containing that object class, resulting in good reconstruction performance. In contrast, the reconstructions in the third column were obtained at the end of CL after all learning exposures had been introduced. Note that the model received only one exposure to each object class. Surprisingly, the quality of the reconstruction produced by the final model slightly improves relative to the first exposure, which is evidence for the lack of negative backward transfer. Fig.~\ref{fig:bwt_fwt}b illustrates FWT performance. While the model was never trained on these unseen classes, the quality of the 3D reconstructions improves steadily as learning progresses, proving strong and surprising evidence for positive FWT and the ability to leverage a shared representation between tasks using only fine-tuning via vanilla SGD. We believe that our novel findings provide crucial insights into the feasibility of systems that require continual learning of object shape.



\begin{table}[t!]
    \begin{center}
\resizebox{0.8\linewidth}{!}{

\begingroup
\setlength{\tabcolsep}{4pt} 
\renewcommand{\arraystretch}{1.2} 

\begin{tabular}{c|c}
 \begin{tabular}{@{}c@{}}Input Rep.$\rightarrow$ Output Rep. \vspace{-3pt}\end{tabular} & \begin{tabular}{@{}c@{}}Reconstruction Tasks \vspace{-3pt}\end{tabular}\\
\hline
2D $\rightarrow$ 3D   & \begin{tabular}{@{}c@{}}Single-view Image 3D Shape Rec. (Figs.~\ref{fig:3Dshape}a,~\ref{fig:3Dshape}b) \vspace{-1pt} \end{tabular}\\
\hline
2.5D $\rightarrow$ 3D  & \begin{tabular}{@{}c@{}}Single-view Depth 3D Shape Rec. (Figs.~\ref{fig:3Dshape}a,~\ref{fig:3Dshape}b) \vspace{-1pt} \end{tabular} \\
\hline
3D $\rightarrow$ 3D & \begin{tabular}{@{}c@{}}Single-object Pointcloud 3D Shape Rec. (Fig.~\ref{fig:3Dshape}c) \vspace{-1pt} \end{tabular} \\
\hline
2D $\rightarrow$ 2.5D &  \begin{tabular}{@{}c@{}}Single-view Depth Pred. (Fig.~\ref{fig:3Dshape}d)\\Single-view Surface Normals Pred. (Fig.~\ref{fig:3Dshape}d) \vspace{-1pt} \end{tabular}\\
\hline
2D $\rightarrow$ 2D & \begin{tabular}{@{}c@{}}Image auto-encoding (Sup.) \\Single-view Silhouette Pred. (Sup.) \vspace{-1pt} \end{tabular}\\

\hline
\end{tabular}


\endgroup
}

\end{center}

    \vspace{-10pt}
    \caption{Summary of the reconstruction tasks we evaluate that demonstrate robustness to catastrophic forgetting. There are 5 types of tasks based on the input to output representation mapping.}

    \label{table:task}
\vspace{-20pt}
\end{table}

In summary, this paper makes the following contributions: 1) Formulation of \emph{continual object shape reconstruction} tasks (Tbl.~\ref{table:task}), including complete 3D shape reconstruction from different input modalities and visible 3D surface (2.5D) reconstruction (Sec.~\ref{sec:problem_formulation}); 2) The surprising finding that these tasks exhibit lack of negative backward transfer and catastrophic forgetting (Sec.~\ref{sec:cl_tasks}); 3) Evidence for improved generalization ability which is indicative of positive forward transfer (Sec.~\ref{sec:forward}); 4) Novel output distribution shift measurement which demonstrates that smaller output distribution shift across learning exposures leads to better knowledge transfer in continual learning (Sec.~\ref{sec:analysis}); 5) Using single-view 3D shape reconstruction as a \emph{proxy task} for classification is effective given a limited exemplar budget (Sec.~\ref{sec:proxy}).

\section{Related Work}\label{sec:rel_work} Our work is most closely-related to four bodies of prior work: 1) CL works outside of the image classification paradigm (relevant to our findings on CL for reconstruction), 2) Analysis of CL (relevant to our output distribution shift analysis), 3) Generalization ability of models for single image 3D shape reconstruction (relevant to our investigation of generalization ability of CL single-view 3D shape reconstruction models), and 4) CL for classification (relevant to our proxy representation task findings).

\noindent\textbf{CL of Non-Classification Tasks}. We are the first to investigate and demonstrate that a set of CL tasks is intrinsically robust to catastrophic forgetting. While most prior CL works have addressed image classification, a few prior works have addressed various other tasks: Aljundi et al.~\cite{Aljundi_2019_CVPR} studied the problem of actor face tracking in video, while~\cite{Michieli_2019_ICCV,Cermelli_2020_CVPR,Maracani_2021_ICCV,douillard2021plop} explored image segmentation. Some works~\cite{shmelkov2017incremental,liu2020multitask,Wang_2021_ICCV} investigated incremental object detection while ~\cite{lesort2019generative,wu2018memory} learned image generation. Elhoseiny et al.~\cite{elhoseiny2018exploring} examined continual fact learning by utilizing a visual-semantic embedding. Wang et al.~\cite{wang2021continual} studied CL of camera localization given an input RGB image while~\cite{Cai_2021_ICCV} explored online CL of geolocalization with natural distribution shift in the input that occurs over real time. Others~\cite{abel2018state,kaplanis2018continual,xu2018reinforced} focused on reinforcement learning. 

Most closely related to our work is Yan et al.~\cite{Yan_2021_ICCV} that investigated continual learning of scene reconstruction. Similar to our work, they employed implicit shape representation (signed-distance-field) to represent 3D scenes. In contrast, this work aimed to continually reconstruct the input scene given a stream of depth images from different views. The input distribution shift in this setting is the shift between one view of the scene to another and the objective is to produce a smooth representation of the same input scene observed over time. Our work on the other hand, explores CL of reconstruction task in the context of visual classes, which is more challenging since the underlying semantics in the inputs change over time. Note that \emph{all of these CL works reported challenges with catastrophic forgetting commensurate with the classification setting.}

\noindent\textbf{Analysis of Continual Learning}. Our analysis of the behavior of CL tasks is most closely related to the body of works that analyzes general dynamics of CL~\cite{knoblauch2020optimal,Verwimp_2021_ICCV}. While~\cite{Verwimp_2021_ICCV} examined the benefits and drawbacks of rehearsal methods in CL,~\cite{knoblauch2020optimal} showed that optimal CL algorithms solve an NP-HARD problem and require the ability to approximate the parameters that optimize all seen tasks. While~\cite{lesort2021understanding} discussed the different concept drifts in CL, our analysis focuses more on the output distribution shift that can be used as a means to understand the knowledge transfer ability of various CL tasks.

\noindent\textbf{Generalization in Batch-Mode 3D Shape Reconstruction}. Our analysis of the generalization ability of CL 3D single-view shape reconstruction task in Sec.~\ref{sec:forward} is based on prior works that investigate the ability of single image 3D shape reconstruction models to generalize to unseen shape categories in batch mode~\cite{thai20203d,genre,shin2018pixels}. We are the first to provide generalization analysis of these models in the CL setting, utilizing the 3-DOF VC approach which was shown to learn a more general shape representation than the object-centered (OC) approach.

\noindent\textbf{CL for Classification}. Our work on a reconstruction-based proxy task for CL classification (Sec.~\ref{sec:proxy}) is unique, but it is peripherally-related to other CL works which explore alternative classification losses or forms of supervision. We share with Yu et al.~\cite{Yu_2020_CVPR} the use of the nearest-class-mean (NCM) classification rule. We use NCM for classification based on a latent shape representation trained without class supervision, while Yu et al. use NCM for classification in an embedding layer which is trained with ground-truth class labels. Another related work by Rao et al.~\cite{rao2019continual} performs unsupervised CL in a multi-task setting where the boundaries between tasks are unknown. In contrast, our unsupervised training paradigm utilizes single-view 3D shape reconstruction as a proxy task.

\section{Problem Formulation}\label{sec:problem_formulation}

\begin{figure*}[t!]
\centering
\includegraphics[width=0.85\linewidth]{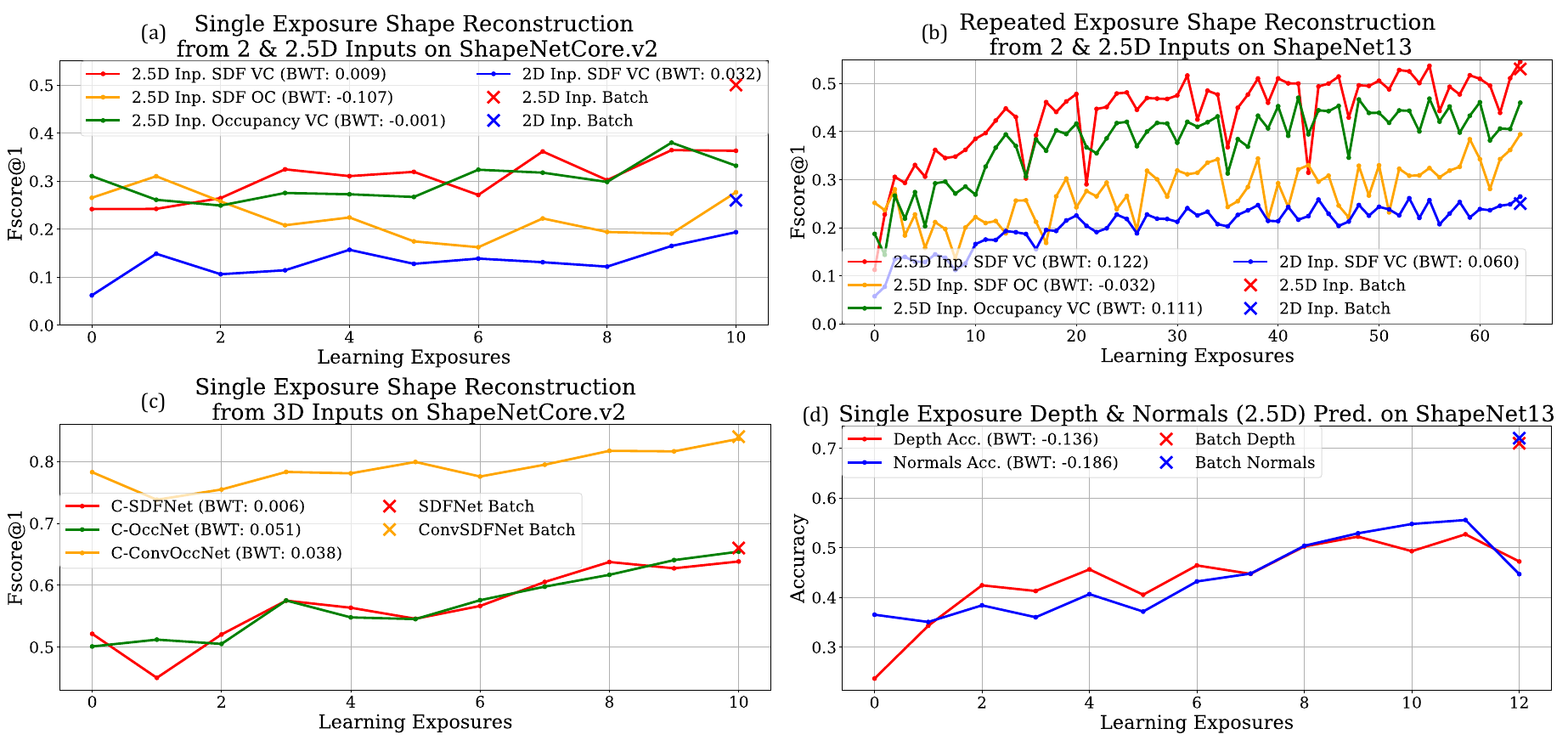}
\vspace{-10pt}
\caption{(a) Performance of shape reconstruction methods with 2D and 2.5D inputs when presented with a single exposure for each category from all 55 categories of ShapeNetCore.v2, 5 classes/exposure
(b) repeated exposures case on ShapeNet13 with 10 repeated exposures, 2 classes/exposure (c) single exposure case on ShapeNetCore.v2 of shape methods with 3D inputs. (d) Results for 2.5D estimation. Performance in terms of thresholding accuracy ($\delta=1.25$) for depth prediction and thresholding cosine distance ($\delta = 0.9$) for surface normals. Backward transfer is reported in parenthesis. Catastrophic forgetting does not happen to any of the algorithms in any case.
}\label{fig:3Dshape}
\vspace{-15pt}
\end{figure*}


\noindent\textbf{Continual Learning of Reconstruction}. At each learning exposure $t$, the learning model observes the data $\{(\boldv{x}_i^{(t)},\boldv{y}_i^{(t)})\}_{i=1}^{N_t}\sim\mathcal{D}_t$ indexed by $t\in\{1,2\dots,T\}$. For example, single-view 3D shape reconstruction aims to output the 3D shape of the object represented in the input image. The model learns to optimize the parameters $\theta_t$ of the function $f_{\theta_t}:\mathcal{X}_t\rightarrow\mathcal{Y}_t$  by minimizing the supervised loss 
$\mathcal{L}(\theta_t) = \mathbb{E}_{\mathcal{D}_t}[\ell(\boldv{y}^{(t)}, f_{\theta_t}(\boldv{x}^{(t)}))]$ where $\ell(\cdot,\cdot)$ is some loss function associated with the specific reconstruction task. 


We employ the notion of \emph{single exposure} to refer to the standard continual learning paradigm where data is introduced sequentially and never revisited while \emph{repeated exposures} refers to the paradigm introduced in \cite{stojanov2019incremental} where data can be revisited after being learned. In this setting, each visual class occurs a fixed number of times (e.g. 10 repetitions) in random order\footnote{Details discussed in the Supp.}. Note that in this work, we assume that each $\mathcal{D}_t$ is defined over a set of $M_t$ visual categories.\footnote{Organizing shapes into categories allows us to characterize how new shape concepts are introduced during learning.}


\noindent\textbf{Training}.  During training, the learning model \emph{does not} have access to previously seen data $\mathcal{D}_{1:t-1}$. We optimize the parameters $\theta_t$ of the function $f$ continuously at each learning exposure upon observing \emph{only} the data stream $\mathcal{D}_t$. Specifically, the learned parameters $\theta_{t-1}$ at exposure $t-1$ serve as the initialization parameters for the model at exposure $t$, which we refer to as continuous representation learning. This is the standard SGD training that has been shown to suffer from catastrophic forgetting in prior works. Without any further heuristics such as additional losses, external memory or other methods employed, this technique is referred to as \emph{fine-tuning strategy}~\cite{li2017learning}.

\noindent\textbf{Evaluation}. At test time we consider the following metrics at each learning exposure: 1) $\text{Acc}_t^s$: accuracy on all known categories (Secs.~\ref{sec:cl_tasks},~\ref{sec:proxy}) and 2) $\text{Acc}_t^g$: accuracy on a fixed, held out set of unseen classes that are never explicitly learned (Sec.~\ref{sec:forward}). Plotting the average accuracy at all learning exposures results in the learning curve of the CL model. All accuracy metrics reported are in range $[0,1]$.

We further report backward and forward transfer metrics~\cite{lopez2017gradient} in addition to the average performance curve at each learning exposure. Specifically, backward transfer (BWT) measures the average change in performance in the last learning exposure w.r.t when the concepts are first introduced and forward transfer (FWT) indicates the average change in performance between the random initialization and the performance of the learning exposure right before the concepts are introduced. Note that while BWT is bounded in $[-1,1]$, FWT depends on the random initialization performance on each dataset. A more successful CL learner will demonstrate higher BWT and FWT.

\section{Single Object Shape Reconstruction Does Not Suffer from Catastrophic Forgetting}\label{sec:cl_tasks}

Tbl.~\ref{table:task} lists the five types of reconstruction tasks that we evaluate in this work, which include 3D, 2.5D, and 2D output domains. Our key finding is that CL tasks from each of these five types do not suffer from catastrophic forgetting. 
It is important to emphasize that the ``continual learning" algorithm used in this section is the simple fine-tuning strategy specified in Sec.~\ref{sec:problem_formulation}, that is known to perform very poorly for  classification tasks. Specifically, \emph{we do not need to utilize additional losses, external memory, or other methods to achieve good continual learning performance}. 

Note that different categories of shapes exhibit significant domain shift that poses significant challenges to continual learning. For example, the categories ``chair" and ``bowl" in ShapeNet define very different 3D data distributions with no parts in common. From this point of view, it is quite surprising that we do not observe forgetting for such continual reconstruction tasks. We therefore organize shapes by category in constructing our learning exposures, so that the category label is a means to characterize the domain shift between successive exposures.

Our findings for learning 3D shape reconstruction and 2.5D prediction are presented in Secs.~\ref{sec:3d-shape-recon} and ~\ref{sec:DN-recon} respectively. We additionally conduct experiments on 2D reconstruction tasks in the Sup. In Sec.~\ref{sec:discussion} we present two possible simple explanations for the lack of catastrophic forgetting and provide empirical evidence that rejects these hypotheses. We report $\text{Acc}_t^s$ as described in Sec.~\ref{sec:problem_formulation} and backward transfer for all the experiments.

\subsection{Single Object 3D Shape Reconstruction} 
\label{sec:3d-shape-recon}
We first present reconstruction tasks where the output representation is in 3D. Specifically, given a single image or sparse pointcloud as the input, the goal of the desired function $f$ is to produce a 3D surface representation of the object present in the input. We focus our analysis on signed-distance-fields (SDF) since it was identified to achieve superior performance in the batch setting~\cite{thai20203d,xu2019disn}. The SDF value of a point in 3D space indicates the distance to the closest surface from that point, with the sign encoding whether the point is inside (negative) or outside (positive) of the watertight object surface. Thus, the 3D surface is represented as a zero-level set where all the points lying on the surface of the object have SDF value 0. 




 \noindent\textbf{Approach}. We utilize SDFNet~\cite{thai20203d} and OccNet\footnote{OccNet utilizes continuous occupancies as the 3D representation.}~\cite{mescheder2019occupancy} as backbone architectures for CL with 2D and 2.5D input representations where inputs are single-view RGB images and ground truth depth and normal maps respectively. We train both methods with the 3-DOF VC representation (varying in azimuth, elevation and camera tilt) from~\cite{thai20203d}, which was shown to give the best generalization performance.\footnote{3-DOF VC SDFNet is the current SOTA for single image 3D shape reconstruction.} We also train with object-centered (OC) representation for SDF representation, in which the model is trained to output the shape in the canonical pose. For 3D input representations where inputs are sparse 3D pointclouds, we further examine a variant of ConvOccNet~\cite{peng2020convolutional} that outputs SDFs instead of continuous occupancies (ConvSDFNet). In the Supp. we additionally show results on a standard pointcloud autoencoder following in~\cite{achlioptas2018learning}.
 
 



\noindent\textbf{Datasets \& Metric}. We train on all 55 classes of ShapeNetCore.v2~\cite{chang2015shapenet} (52K instances) with 5 classes per exposure for the single exposure case, and on the largest 13 classes of ShapeNetCore.v2 (40K meshes), denoted as ShapeNet13, with 2 classes per exposure for the repeated exposure case. 
Note that ShapeNetCore.v2 is currently the largest shape dataset with category labels and ShapeNet13 is the standard split for 3D shape reconstruction. Each exposure is generated from all of the samples from the training split of each category currently present.\footnote{For instance, if chair and table are present in the current learning exposure, the model will be trained on all chairs and tables in the respective training splits.} Following prior works in shape reconstruction~\cite{thai20203d,xu2019disn,tatarchenko2019single} we report the average FS@1 at each learning exposure. We use SDFNet as the batch reference for 2D and 2.5D inputs. For 3D inputs we include ConvSDFNet batch performance. All models are trained from random initialization.

\begin{figure*}
\begin{minipage}{0.49\linewidth}
\centering
\includegraphics[width=0.9\linewidth]{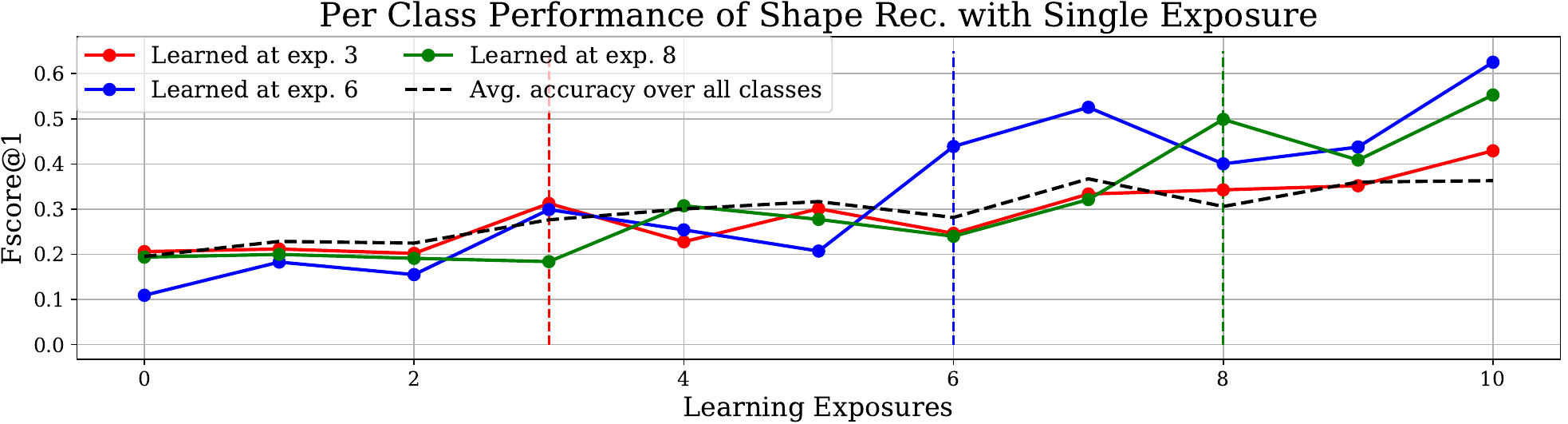}
\vspace{-10pt}
\caption{Reconstruction performance of three classes throughout all learning exposures. Performance is consistently poor before a class is learned, but significantly improves after and remains high in subsequent exposures. Dotted back line indicates the average accuracy over all classes (learned and as-yet unseen).}
\label{fig:per_cls}
\end{minipage}
\hspace{7pt}
\begin{minipage}{0.49\linewidth}
\centering
\includegraphics[width=0.9\linewidth]{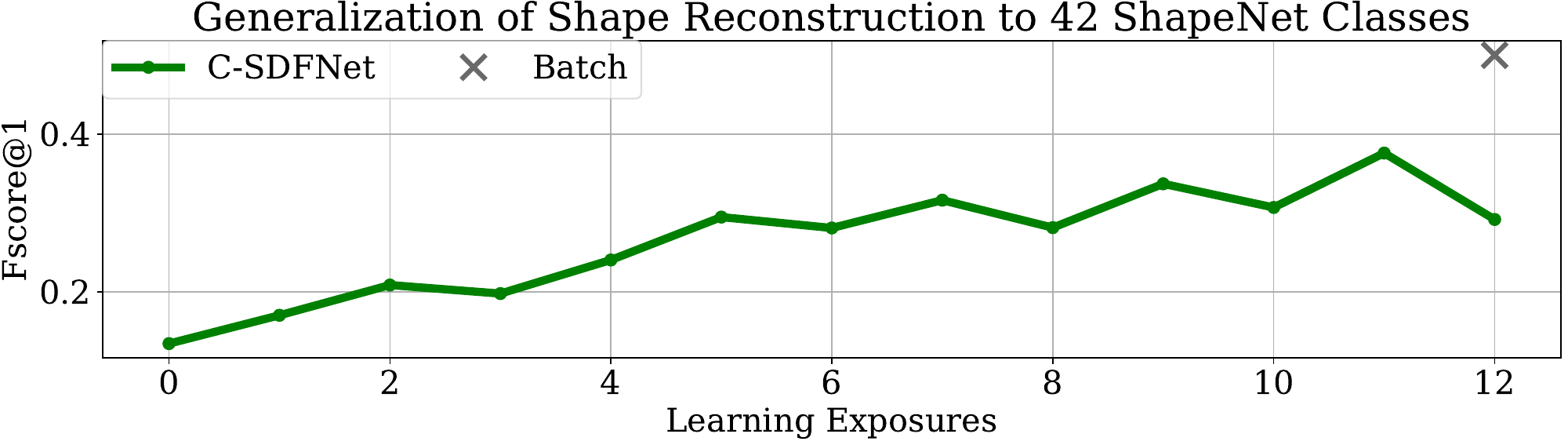}
\vspace{-10pt}
\caption{Generalization performance to 42 unseen categories of ShapeNetCore.v2 of continual SDFNet (C-SDFNet). Generalization ability increases over time demonstrates the benefit of continuous representation learning. Note that all C-SDFNet is trained with 3-DOF VC approach and 2.5D input.}
\label{fig:forward_transfer}
\end{minipage}
\vspace{-20pt}
\end{figure*}

\noindent\textbf{Results}. The results are shown in Figs.~\ref{fig:3Dshape}a,~\ref{fig:3Dshape}b and~\ref{fig:3Dshape}c for single and repeated exposures on all single object 3D shape reconstruction settings (last 3 rows of Tbl.~\ref{table:task}). For single exposure with 2D and 2.5D inputs (Fig.~\ref{fig:3Dshape}a), all algorithms maintain their accuracy over time and even exhibit a slight upward trend of increasing accuracy while for 3D inputs (Fig.~\ref{fig:3Dshape}c) the performance increases more consistently over time and is on par with batch. Note that we conducted 3 runs and the results converge to the same conclusion with an average std of 0.02 at each learning exposure. All models including the model trained with OC representation \emph{do not suffer from catastrophic forgetting} as evidenced by the minimal negative and even positive backward transfer. This is surprising since we are not taking any steps to ameliorate catastrophic forgetting and each learning exposure presents a significant domain shift, as the learner must incorporate information about the shape of a new object class. Since our findings hold on various model architectures with different input/output representations, \emph{this possibly reflects a basic property of the shape reconstruction problem} rather than the inductive biases of a particular model. 


In the repeated exposures setting (Fig.~\ref{fig:3Dshape}b), the performance of both SDFNet and OccNet when trained with 3-DOF VC \emph{improves significantly over time}, and eventually performs on par with batch.\footnote{The 65 learning exposures (x axis) in Fig.~\ref{fig:3Dshape}b result from 13 ShapeNet classes divided by 2 classes per exposure with 10 repetitions each.} These models achieve significant positive BWT which indicates that catastrophic forgetting is mitigated. Unlike the experiments in~\cite{stojanov2019incremental}, which showed similar asymptotic behavior for classification accuracy, these results were obtained \emph{without}  exemplar memory or other heuristics. Note that SDFNet trained with OC does not show a significant increase as 3-DOF VC over time. This complements the finding in~\cite{thai20203d} that training with 3-DOF VC results in a more robust feature representation.

\subsection{Single-view 2.5D Sketch Prediction}
\label{sec:DN-recon}
The task in Sec.~\ref{sec:3d-shape-recon} requires the model to infer the global 3D structure of each object. In this section we investigate the related task of estimating depth and surface normals (2.5D) from RGB input images in the single exposure case (Tbl.~\ref{table:task}, second row). We adopt the U-ResNet18-based  MarrNet~\cite{wu2017marrnet} architecture, with an ILSVRC-2014~\cite{russakovsky2015imagenet} pre-trained ResNet18 for the image encoder. We evaluate depth prediction using the commonly used thresholding accuracy~\cite{koch2018evaluation,Ramamonjisoa_2019_ICCV}. For normals prediction, we report the accuracy based on the cosine distance threshold between the predicted and ground truth surface normals~\cite{wang2020surface}\footnote{More details in Supp.}. Fig.~\ref{fig:3Dshape}d demonstrates that single exposure 2.5D prediction does not suffer catastrophic forgetting as the accuracy increases over time. These findings further extend the 3D shape reconstruction results. While the performance of some CL models learned with single exposure when all data has been seen does not reach batch (for 2D$\rightarrow$3D, 2.5D$\rightarrow$3D, and 2D$\rightarrow$2.5D tasks), we note that these tasks are sufficiently challenging (even in the batch setting where data is iid) and emphasize that the surprising positive trend of the curves has never been shown in prior CL works. 

We conduct additional experiments on continual 2D to 2D mapping that includes learning to segment foreground/background given an RGB input image and image autoencoding. We refer to the Supp. for details.

\subsection{Discussion of CL Object Shape Reconstruction}\label{sec:discussion}

We have identified (for the first time) a set of \emph{continual object shape reconstruction} tasks that do not suffer from catastrophic forgetting (see Fig.~\ref{fig:3Dshape}) when models are trained using \emph{standard SGD} without any heuristics. A key question is why this is happening. We examine two possible simple explanations for the CL performance of single-view 3D shape reconstruction: 1) The learner encodes ``low-level" features of the inputs that are present for all object classes and facilitate easy generalization, and 2) the domain shift between consecutive learning exposures is small, making the CL problem almost trivial. We find that neither of these hypotheses is supported by our findings, suggesting that the behavior we have discovered is nontrivial, which can motivate for future research and investigation.


\noindent\textbf{Low-level Features}. Are there some low-level visual properties shared by all 3D shapes that the learner can index on to solve CL? This seems implausible, as single image reconstruction is a challenging task that requires learning mid- to high-level properties of classes of shapes (e.g., concavities in bowls and tubs, protrusions in chairs and tables) in order to learn to reconstruct occluded surfaces. 
Since shape reconstruction losses penalize the entire 3D shape (including occluded surfaces), good performance on unseen classes requires nontrivial generalization. We also demonstrate that learned shape representations encode categorical information: We fit a linear classifier on top of the shape features extracted from SDFNet trained on ShapeNetCore.v2 (all 55 classes) and we find that it obtains 65\% accuracy, compared to 16\% for random features and 42\% for ImageNet pretrained features. This shows that the learner is encoding complex properties of 3D shape in solving the task.

\noindent\textbf{Domain Shift}. In Fig.~\ref{fig:per_cls}, we present quantitative evidence that continual shape reconstruction is characterized by significant class-based domain shift: The per-class reconstruction performance for three representative classes is low before each class is learned (introduced in the training data) and then rises significantly after. It's clear that the learned representation is responding to the properties of each class, and yet there is very little forgetting. 
We present additional analysis of domain shift in Sec.~\ref{sec:analysis}, to shed further light on this phenomenon. In summary, we argue that CL object shape reconstruction is solving a nontrivial task which requires a complex generalization ability, and therefore merits further investigations in future work using the framework we have provided.




\begin{figure*}
\vspace{-15pt}
\centering



\begin{minipage}{0.45\textwidth}
\centering
    \includegraphics[width=0.9\linewidth]{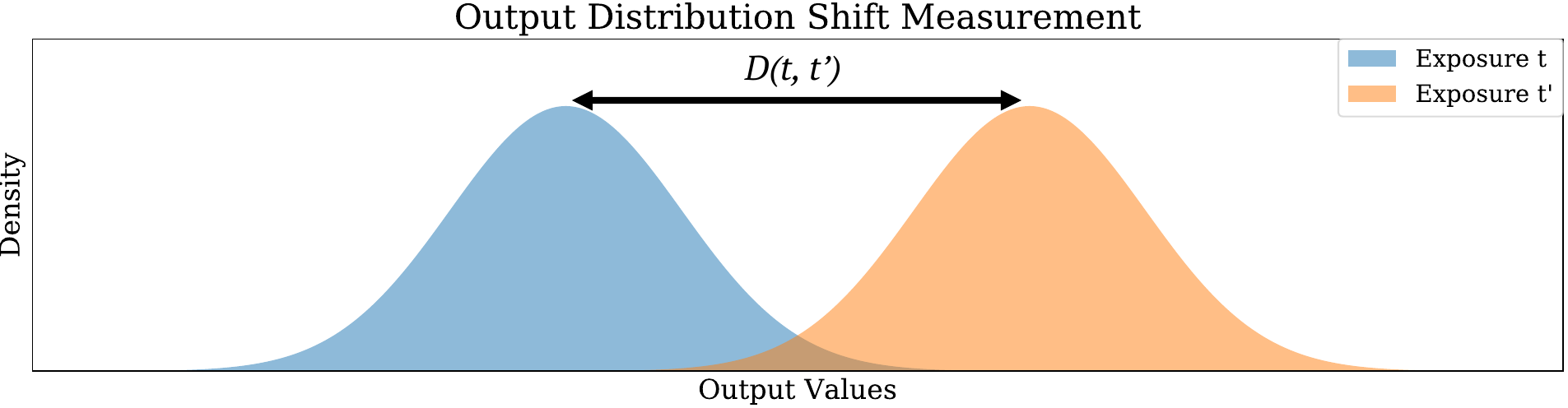}
    \vspace{-10pt}
    \caption{Visualization of our output distribution shift measurement. We use the first Wasserstein distance metric (EMD) which measures the least amount of required work to transform one distribution into another to obtain the output distribution shift between different exposures $t$ and $t'$.}
    \label{fig:emd}
\end{minipage}
\hspace{7pt}
\begin{minipage}{0.5\textwidth}
\centering
\captionsetup{type=table} 
\resizebox{0.9\linewidth}{!}{

\begingroup

\setlength{\tabcolsep}{4pt} 
\renewcommand{\arraystretch}{1.2} 

\begin{tabular}{c|c|c|c}
 Task  & Normalized Mean Dist. $\downarrow$ & BWT $\uparrow$ & FWT $\uparrow$ \\
\hline
Sil Pred. & 0.075 & -0.003 & 0.836\\
VC 3D Shape Rec. & 0.076 & -0.123 &    0.105   \\    
Depth Pred. & 0.084 & -0.136 & 0.094\\
OC 3D Shape Rec. & 0.105 & -0.220 & 0.090\\
\hline
\end{tabular}
\endgroup
}
    \caption{The relationship between the mean output distribution distance across learning exposures and BWT and FWT, with lower distance leading to better knowledge transfer (higher BWT and FWT).}
    \label{table:dist}
\end{minipage}
\vspace{-20pt}
\end{figure*}

\section{Generalization of CL 3D Reconstruction}\label{sec:forward}
In this section, we discuss the ability of the learning model to propagate useful representations learned in the past to current and future learning exposures (FWT). We focus our analysis on the challenging problem of single-view 3D shape reconstruction. While generalization to unseen classes has been studied extensively in the batch setting of single-view 3D shape reconstruction, and has been identified to be a significantly challenging problem~\cite{genre,thai20203d}, we are the first to analyze this behavior in a continual learning setting. In this section, we report $\text{Acc}_t^g$.

We conduct our experiments on ShapeNet13 with single exposure and 1 shape class per learning exposure on continual SDFNet (C-SDFNet) (Sec.~\ref{sec:3d-shape-recon}). We evaluate C-SDFNet on a held out set of  42 classes of ShapeNetCore.v2 with 50 instances for each category (Fig.~\ref{fig:forward_transfer}). The model performs poorly on the unseen classes after the initial learning exposures, which demonstrates that it is significantly challenging to generalize to novel categories after learning on only a few classes. However, the performance improves over time as more classes are learned. This illustrates benefit of continuous representation learning as a useful feature that aids generalization and improves the performance on novel classes over time. In Fig.~\ref{fig:bwt_fwt} we show qualitative results that demonstrate positive knowledge transfer ability of single-view 3D shape reconstruction task. 

In the Supp. we provide further evidence that continuous representation training is beneficial for CL of single-image 3D shape reconstruction by comparing with an episodic training approach that was shown to achieve competitive performance in CL classification. We additionally present a simple yet competitive CL classification baseline that employs continuous representation update strategy.

\section{Analysis of Knowledge Transfer Ability}\label{sec:analysis}
\label{subsec:hyp}
Our findings in Secs.~\ref{sec:cl_tasks} and~\ref{sec:forward} have highlighted the significance of knowledge transfer in CL reconstruction. While BWT and FWT quantify the knowledge transfer during CL, they require training and evaluating computationally expensive CL models.\footnote{Training and evaluating 3D shape reconstruction from 3D inputs on ShapeNetCore.v2 takes 3 days on two NVIDIA GeForce RTX 2080Ti GPUs. On the other hand, computing output distribution distance only takes $\approx45$ minutes which is two orders of magnitude more efficient} Furthermore, these measures only reflect the performance of specific CL algorithms and do not speak to a CL task in general. In this section, we attempt to gain more insight into knowledge transfer given a task and a dataset in an \emph{algorithm-agnostic} manner, by focusing on changes in the output distribution. We use this approach to further analyze the benefit of exemplar memory in classification (see details in Supp.). We first state the hypothesis connecting the output distribution to CL task knowledge transfer ability.






\noindent\textbf{Hypothesis:} When  the distance of the output distribution between each learning exposure becomes smaller, backward and forward transfer increase for any CL method. 

\vspace*{0.5em} We now present the intuition behind our formulation. Let $\mathcal{D}$ be some dataset consisting of two parts $\mathcal{D}_1$ and $\mathcal{D}_2$ that are independently generated. During batch training we optimize the parameters $\theta\in\Gamma$ where $\Gamma$ is the model parameter space by minimizing the negative likelihood. 
Since $\mathcal{D}=\mathcal{D}_1\cup\mathcal{D}_2$ and $\mathcal{D}_1$ and $\mathcal{D}_2$ are independent, the negative likelihood reduces to $-\log p(\mathcal{D}_1\vert\theta)-\log p(\mathcal{D}_2\vert\theta)=-\log p(Y_1\vert X_1,\theta)-\log p(Y_2\vert X_2,\theta)$ where $X_t \sim \mathcal{X}$ and $Y_t \sim \mathcal{Y}$ are the inputs and outputs respectively.
During continual learning when $\mathcal{D}_1$ and $\mathcal{D}_2$ are learned sequentially, we optimize $\mathcal{L}_1(\theta_1)=-\log p(\mathcal{D}_1\vert\theta_1)$ and  $\mathcal{L}_2(\theta_2)=-\log p(\mathcal{D}_2\vert\theta_2)$ separately where $\theta_1, \theta_2\in\Gamma$ are model parameters, which leads to a suboptimal solution for $\mathcal{L}(\theta)$. When the distance between the conditional distributions $Y_1\vert X_1$ and $Y_2\vert X_2$ is small, it is more likely that the optimal parameters $\theta_1$ for $\mathcal{L}_1$ coincides with the optimal parameters $\theta_2$ for $\mathcal{L}_2$ and hence the joint parameters $\theta$ that optimize the batch training model.

\noindent\textbf{Analysis}.
We now demonstrate the empirical evidence for the earlier hypothesis. Note that in all of the following analyses, the input $X_t$ is defined to be a visual object category.

\emph{Distribution Distance Metric}. We use the first Wasserstein distance metric (EMD) to quantify the distance between two output distributions. EMD was introduced by Rubner et al.~\cite{rubner2000earth} to measure the structural similarity between distributions. In contrast to other statistical measurements like KL divergence or Chi-squared statistics, EMD can be used to measure the similarity between both continuous and discrete distributions with different supports.
Given distributions $u$ and $v$, we define {$d(u,v)=\inf_{\pi\in\Gamma(u,v)}\int_{\mathbb{R}\times\mathbb{R}}\vert x-y\vert d\pi(x,y)$}
and express the distance between two learning exposures $t$ and $t'$ as
\begin{equation}\label{eq:dist}
\small
    D(t,t')=\frac{1}{\vert\mathcal{S}\vert} \int_{s\in\mathcal{S}}d(u_t,u_{t'})\text{d}s
\end{equation}
where $u_t$ and $u_{t'}$ are the output distributions at exposures $t$ and $t'$ respectively and $\mathcal{S}$ is the support set of $u_t$ and $u_{t'}$ (please see Fig.~\ref{fig:emd} for a visual illustration). We now analyze the output distribution shift for different CL tasks. Note that we normalize the distribution shift by the range of the output values so that they are defined over a support set of the same length.

\begin{figure*}[t!]
\begin{minipage}[t!]{\linewidth}
\centering
\vspace{-20pt}
    \includegraphics[width=0.8\linewidth]{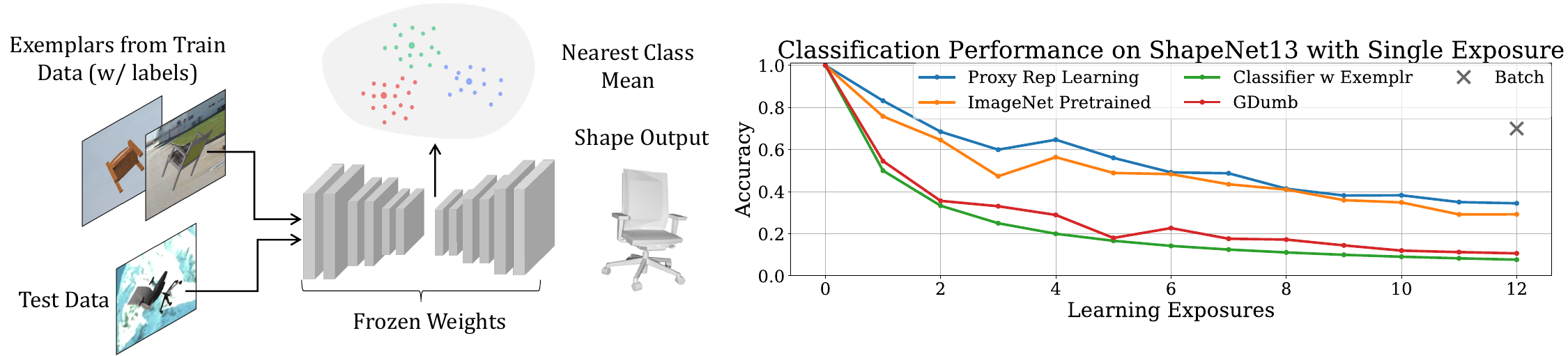}
    \vspace{-10pt}
    \caption{(a) Our approach to CL classification using reconstruction as a proxy task: We extract the feature representations of the exemplars and test data via a forward pass on the trained 3D reconstruction model. Classification is done via Nearest Class Mean. (b) CL performance (Proxy Rep Learning) is shown for ShapeNet13 with RGB input. Given a limited exemplar budget, we outperform ImageNet pretrained features and classification baselines.}
    \label{fig:exp_proxy}
\end{minipage}%
\vspace{-20pt}
\end{figure*}



\emph{3D Shape Reconstruction}. In this setting, the output $Y_t^{\text{SDF}}$ represents the ground truth SDF values for the support set $\mathcal{S}$ consisting of 3D coordinates. We first select $1000$ 3D points uniformly in a unit grid of resolution $128^3$. For each shape class, we randomly sample $1000$ objects. Each 3D point $q_i$ defines a distribution of SDF values within a shape class $P_{q_i}^{(t)}=\mathbb{P}(Y_t^{\text{SDF}}\vert q_i, X_t)$. From Eq.~\ref{eq:dist}, the final output distribution distance between each shape class is {$D(t,t')=\frac{1}{N_q}\sum_{i=1}^{N_q}d(P_{q_i}^{(t)}, P_{q_i}^{(t')})$} where $N_q$ is the number of 3D points. We present the results for both OC and 3-DOF VC representations described in Sec.~\ref{sec:3d-shape-recon}.

\emph{2.5D Depth Prediction and 2D Silhouette Prediction}. In this setting, $Y_t^{\text{pix}}$ represents the value of each pixel of the input $X_t$ (depth value and binary value for depth and sihouette pred. respectively). The support set $\mathcal{S}$ is the set of 2D pixel coordinates. Each pixel $p_i$ then defines a distribution of pixel values within a class $P_{p_i}^{(t)}=\mathbb{P}(Y_t^{\text{pix}}\vert p_i, X_t)$. The output distribution distance between each class is {$D(t,t')=\frac{1}{N_p}\sum_{i=1}^{N_p}d(P_{p_i}^{(t)}, P_{p_i}^{(t')})$} where $N_p$ is the number of pixels. For depth prediction, we first center crop the input images. For each class we randomly sample $800$ objects and for each image sample $1000$ pixels uniformly.


 We first compute the output distribution distance as described above for each task and compare it with the resulting BWT and FWT. To verify the effectiveness of the proposed method and to ensure fairness we continually train each task using the fine-tuning strategy on ShapeNet13 from 2D RGB input images with 1 class per learning exposure and report the average output distribution distance and the BWT and FWT metrics. Tbl.~\ref{table:dist} shows that our hypothesis holds as the small output distribution distance is associated with higher BWT and FWT. This finding further explains the behavior we observed in Figs.~\ref{fig:3Dshape}a,b where VC consistently outperforms OC 3D shape model.
 
 In the Supp., we conduct further investigation on the forgetting phenomenon for CL classification. 

\section{Proxy Task for Continual Classification}\label{sec:proxy}

The robustness of representation learning and the ability to transfer knowledge between learning exposures in single-view 3D shape reconstruction begs the question of whether it could be used as a proxy task for class-IL classification~\cite{van2019three}\footnote{In this setting, the learning model is required to discriminate all learned classes.}. We test that hypothesis here via a simple approach: 
We train a 3D reconstruction model, SDFNet VC on RGB images continually as in Sec.~\ref{sec:3d-shape-recon}, and at inference time we extract the feature from its image encoder with a forward pass. We maintain an exemplar set of $20$ images/class with class labels randomly sampled from the training dataset We do not use the labels for training. Instead, we use the extracted representation to do nearest-class-mean (NCM) classification with the exemplars at testing time. Specifically, the mean feature of each class is first computed from the exemplar set. Then test samples are assigned the label of the closest mean feature via cosine distance (Fig.~\ref{fig:exp_proxy}a). We decide to utilize NCM as a classifier instead of training a fully-connected layer with cross-entropy loss, due to the fact that the exemplar set size is small ($<1$\% of the training data) and it has been shown that linear classifier trained with CE loss tends to overfit significantly when the dataset is imbalanced~\cite{wu2019large,castro2018end}.


We conduct experiments with ShapeNet13 with one class per exposure. We first show that the feature representation learned by the single-view 3D shape reconstruction task is discriminative despite not having access to ground truth labels during training. We compare the performance of the proxy classifier against an ImageNet pretrained feature representation model. Specifically, we extract the feature from the ImageNet pretrained ResNet18 via a forward pass and use NCM as the classifier with the same exemplar set size as the proxy classifier. Fig.~\ref{fig:exp_proxy}b shows evidence that shape features are more beneficial for continual classification than the rich discriminative feature representation from ImageNet. We further compare the proxy classifier against two classification baselines: GDumb~\cite{prabhugdumb} and a standard classifier trained continually with cross entropy loss and the same exemplar set, denoted as Classifier with Exemplars. Fig.~\ref{fig:exp_proxy}b shows that the 3D shape proxy classifier outperforms the GDumb and Classifier with Exemplars on ShapeNet13. This demonstrates that a significant amount of discriminative information is encoded in the continual shape representation and suggests that it may be beneficial to explore other proxy tasks as a means to improve CL classification. Note that our goal in this section is to show that the unsupervised pretrained shape features give surprisingly high performance despite not being trained to perform classification or use any heuristics. Therefore, we do not compare our approach extensively to existing SOTA CL classification methods and do not attempt to make SOTA claims against these methods.

\section{Conclusion}
We have identified that CL object 3D shape reconstruction from various modalities exhibit lack of negative backward transfer. In addition, we show that the challenging single-view 3D shape reconstruction task exhibits positive knowledge transfer by investigating the generalization ability of single-view 3D shape reconstruction models in the context of CL for the first time. As a means to characterize the knowledge transfer performance of CL tasks, we provide a novel algorithm-agnostic approach that analyzes output distribution shift. We show that reduction in shift is associated with increased knowledge transfer. We further demonstrate that single-view 3D shape reconstruction task can serve as a promising proxy task for CL classification. We hope that our findings will encourage the community to investigate the intriguing phenomenon observed in CL object shape reconstruction tasks.

\section{Acknowledgement}
We would like to thank Miao Liu, Meera Hahn, and Maxwell A. Xu for the helpful discussion. This work was supported by NIH R01-MH114999 and NSF Award 1936970. This paper is dedicated to the memory of Chengming (Julian) Gu.

{\small
\bibliographystyle{ieee_fullname}
\bibliography{egbib}
}
\clearpage
\appendix
This supplementary material document is structured as follows: In Sec.~\ref{sec:datasets} we describe the training data in more detail; In Sec.~\ref{sec:fwt_sup} we demonstrate the benefit of continuous representation training for CL 3D shape reconstruction; In Sec.~\ref{sec:algorithms} we provide details on the CL algorithms used in the paper, their training implementation details, evaluation metrics and further qualitative results for continual object 3D shape reconstruction task; In Sec.~\ref{sec:exemplar_output}, we provide additional analysis on the effect of exemplar set on the output distribution shift in CL classification; In Section~\ref{sec:rep_exp_expl} we further explain the repeated exposures setting; In Section~\ref{sec:yass} we introduce a simple CL classification algorithm that surprisingly achieves competitive performance with other baselines that employ more complex training heuristics; In Section~\ref{sec:dyrt} we examine the dynamics of the feature representations learned by CL classification algorithms.

\section{Datasets}\label{sec:datasets}
\subsection{ShapeNetCore.v2}
\noindent\textbf{Datasets}: ShapeNetCore.v2 consists of 55 categories with 52K CAD models. This is the current largest 3D shape dataset with category labels. Many prior works in 3D shape reconstruction~\cite{mescheder2019occupancy,choy20163d} utilized a subset of 13 largest categories---ShapeNet13, which consists of approximately 40K 3D instances. Tbl.~\ref{table:shapenet13_stat} lists the 13 categories and the number of samples in each category. For ShapeNet13, we use the standard train/val/test split from prior shape reconstruction works~\cite{choy20163d,mescheder2019occupancy}. We sample 100 objects/category from the test split for evaluation in the repeated exposures case. For the remaining 42 classes in ShapeNetCore.v2, we split randomly with proportion 0.7/0.1/0.2 for train/val/test splits. In the single exposure case on all classes of ShapeNetCore.v2, we randomly sample 30 objects/category for testing. For evaluating novel category generalization ability, we sample 50 objects from the 42 classes.


\noindent\textbf{Rendering}: We render 25 views of RGB images, ground truth silhouette, depth and surface normal maps with resolution $256\times 256$ for each object. Following~\cite{thai20203d}, we generate data using Cycles ray-tracing engine in Blender~\cite{blender} with 3 degree-of-freedom, varying camera azimuth $\theta\in[0,360^\circ]$, elevation $\phi\in[-50^\circ,50^\circ]$ and tilt. For experiments with RGB images as inputs, we render with varying light, specular surface reflectance and random backgrounds from SUN Scenes~\cite{xiao2010sun}.  

\noindent\textbf{SDF Point Sampling Strategy}: For 3D shape reconstruction, training 3D points are sampled more densely close to the surface of the mesh. Following~\cite{thai20203d}, we sample half of the training points within a distance of 0.03 to the surface, $30\%$ with distance in the range $[0.03, 0.1]$ and $20\%$ in the range $[0.1,1.1]$. To train and evaluate OccNet, we obtain mesh occupancy values by binary masking $\mathds{1}\{sdf \leq i\}$ where $i$ is the isosurface value.

\begin{table}[!ht]
    \centering
    \begin{tabular}{|c|c|c|}
    \hline
    ID & Name & Num samples\\
    \hline
    02691156 & airplane & 4045\\
    02828884 & bench & 1813\\
    02933112 & cabinet & 1571\\
    02958343 & car & 3532\\
    03001627 & chair & 6778\\
    03211117 & display & 1093\\
    03636649 & lamp & 2318\\
    03691459 & loudspeaker & 1597\\
    04090263 & rifle & 2373\\
    04256520 & sofa & 3173\\
    04379243 & table & 8436\\
    04401088 & telephone & 1089\\
    04530566 & watercraft & 1939\\
    \hline
    & \textbf{Total} & 39,757\\
    \hline
    \end{tabular}
    \vspace{5pt}
    \caption{Statistics of ShapeNet13.}
    \label{table:shapenet13_stat}
\end{table}

\subsection{CIFAR-100}
This is a standard image dataset consisting of 100 categories with 500 training and 100 testing samples for each category.




\section{Continuous Representation Update Is Effective For CL}\label{sec:fwt_sup}

In this section, we discuss the ability of the learning model to propagate useful representations learned in the past to current and future learning exposures (FWT) and improve performance on learned classes (BWT). We focus our analysis on the challenging problem of single-view 3D shape reconstruction. We first demonstrate that \emph{continuous representation} learning is beneficial as we observe significantly stronger performance compared to \emph{episodic representation} learning for this task. We further note that positive knowledge transfer is obtained, as evidenced by the accuracy improvement on seen and as-yet unseen classes over time. In this section, we report $\text{Acc}_t^f$ and $\text{Acc}_t^g$ in addition to $\text{Acc}_t^s$ (Sec.~3 of main text).

GDumb~\cite{prabhugdumb} is an \emph{episodic representation} learner, designed to test the hypothesis that there is no value in continuous representation learning. Specifically, at each learning exposure, the model is randomly reinitialized and trained from scratch on the exemplar set which ensures that a subset of data from all previous learning exposures is available. This approach surprisingly achieves competitive performance at classification. We hypothesize that in contrast to this observation, continuous representation learning improves the performance in single-view 3D shape reconstruction due to the feasibility of knowledge transfer. In order to test this, we design \emph{GSmart}, an algorithm that continuously trains the feature representation instead of reinitializing the weights at each learning exposure as in GDumb.

\begin{figure*}[t!]
\includegraphics[width=\linewidth]{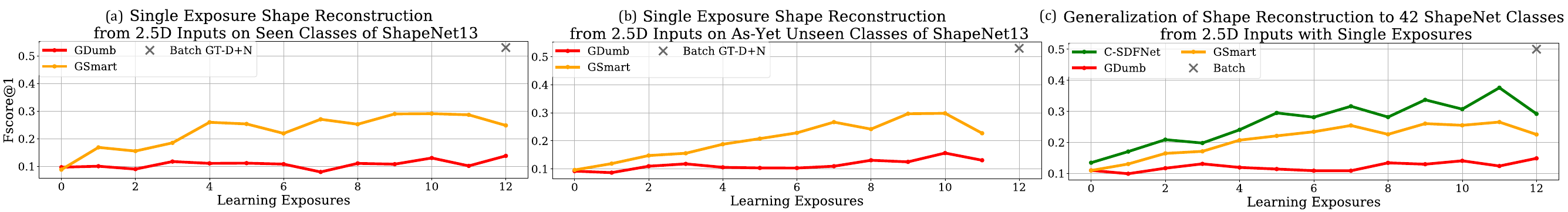}
\vspace{-20pt}
\caption{Performance on (a) seen classes,~(b) as-yet unseen classes of GDumb and GSmart in 3D shape reconstruction with 2.5D inputs on ShapeNet13 with 1K exemplars (3.7\% of training data). GSmart outperforms GDumb by a significant margin. (c) Generalization performance to 42 unseen categories of ShapeNetCore.v2 of GDumb, GSmart and C-SDFNet. Generalization ability of GSmart and C-SDFNet increases over time while constantly staying low for GDumb demonstrates the benefit of continuous representation learning. Note that all models are trained with 3-DOF VC approach.}\label{fig:fwt_sup}
\end{figure*}

We conduct our experiments on ShapeNet13 with single exposure and 1 shape class per learning exposure. We choose $K=1000$ (3.7\% of total training data) to be the exemplar set size and evaluate the performance of the models on all learned classes (Sec.~3 in the main text). In Figs.~\ref{fig:fwt_sup}a, b we observe that the performance of GSmart improves over time and eventually exceeds that of GDumb by $0.15$ FS@1. This significant gap highlights the benefit of continuous representation learning across learning exposure.

\noindent\textbf{Generalization Ability}. We further investigate the ability of single-view 3D shape reconstruction models to generalize to novel categories. We evaluate GDumb, GSmart and C-SDFNet on a held out set of  42 classes of ShapeNetCore.v2 with 50 instances for each category (Fig.~\ref{fig:fwt_sup}c). All algorithms perform poorly on the unseen classes after the initial learning exposures, which demonstrates that it is significantly challenging to generalize to novel categories after learning on only a few classes. However, the performance of C-SDFNet and GSmart improves over time as more classes are learned while GDumb remains low. This illustrates benefit of continuous representation learning as a useful feature that aids generalization and improves the performance on novel classes over time.

\section{Description of Algorithms}\label{sec:algorithms}
\subsection{Single Object 3D Shape Reconstruction}
\noindent\textbf{Architecture}: We adapt SDFNet~\cite{thai20203d} and OccNet~\cite{mescheder2019occupancy} with ResNet-18 encoder for continual training with 2D and 2.5D inputs and SDFNet with PointNet~\cite{qi2017pointnet} encoder for 3D input. Specifically, the architecture consists of an encoder initialized with random weights and a point module which are multiple blocks of fully-connected layers with ReLU activation. Conditional Batch Normalization is used as applying an affine transformation on the output of the point module, conditioned on the feature vector produced by the encoder. 

We additionally adapt a variant of ConvOccNet~\cite{peng2020convolutional}, ConvSDFNet for 3D input where the output representation is SDF instead of continuous occupancies. The pointcloud input is first fed through the PointNet~\cite{qi2017pointnet} encoder to obtain features. These features are then projected onto the $xz$ plane with average pooling and processed by a 2D U-Net~\cite{ronneberger2015u}. Given a query point, bilinear interpolation is used to retrieve the feature associated with that point conditioned on the input. The point module takes the 3D coordinate of the queried point and the associated feature and outputs the SDF value of that point.

To demonstrate that our findings hold on 3D output representations other than implicit continuous representations like occupancies or SDFs, we further conduct experiments on a standard pointcloud autoencoder following in~\cite{achlioptas2018learning}. Specifically, we first extract the features from the pointcloud input using PointNet~\cite{qi2017pointnet} encoder. The decoder is implemented as a stack of linear layers (with dimensions $[512, 1024, 1024]$) with ReLU activations in between. The model outputs a set of 3D points that represents the surface of the input shape.

\emph{GDumb For CL 3D Shape}. We employ SDFNet with ResNet-18 encoder as the backbone architecture and follow the training procedure of GDumb for classification task~\cite{prabhugdumb}. Specifically, we randomly select an exemplar set of size $K=1000$ ($\approx$ 3.7\% of the training data), equally divided for all the seen categories at each learning exposure. We initialize the learning model randomly to train from scratch on the selected exemplar set at each learning exposure. 

\emph{GSmart}. Different from GDumb for CL 3D shape, we continuously update the representation at each learning exposure. Please see Algs.~1,2,3 for the pseudo code of the described CL algorithms.

\begin{algorithm}[t!]

\caption{GDumb for CL 3D Shape}
\SetAlgoLined
\KwIn{Batch training procedure SDFNet$(\theta,\mathcal{D}^\text{train},\mathcal{D}^\text{val})$ that returns the trained parameters $\theta$ and the performance of the trained model on $\mathcal{D}^\text{val}$} 
\KwData{(RGB image, 3D coordinates, SDF values) pair datasets $\mathcal{D}^{\text{train}}=\cup_{i=1}^T\mathcal{D}^{\text{train}}_i$,  $\mathcal{D}^{\text{val}}=\cup_{i=1}^T\mathcal{D}^{\text{val}}_i$}
\KwDef{$\ell:$ weighted $L_1$ loss }
\Init{}{
Exemplar set: $\mathcal{C}=\{\}$
}
\ForEach{\text{learning exposure t in $1,2,\dots,T$} }{
    $\theta\leftarrow\text{RANDOM\_INIT}(\theta)$\\
    $\mathcal{C}\leftarrow$ SELECT\_RANDOM$(\mathcal{C}\cup\mathcal{D}^{\text{train}}_t)$\\
    $\theta_t, \text{acc}_t\leftarrow$ SDFNet$(\theta,\mathcal{C},\mathcal{D}^{\text{val}}_t)$\\
 } 
\KwResult{($\text{acc}_1,\text{acc}_2,\dots\text{acc}_T$)}
\label{algo:gdumb}
\end{algorithm}

\begin{algorithm}[t!]
\caption{GSmart}
\SetAlgoLined
\KwIn{Batch training procedure SDFNet$(\theta,\mathcal{D}^\text{train},\mathcal{D}^\text{val})$ that returns the trained parameters $\theta$ and the performance of the trained model on $\mathcal{D}^\text{val}$}
\KwData{(RGB image, 3D coordinates, SDF values) pair datasets $\mathcal{D}^{\text{train}}=\cup_{i=1}^T\mathcal{D}^{\text{train}}_i$,  $\mathcal{D}^{\text{val}}=\cup_{i=1}^T\mathcal{D}^{\text{val}}_i$}
\KwDef{$\ell:$ weighted $L_1$ loss }
\Init{}{
Exemplar set: $\mathcal{C}=\{\}$
}
\ForEach{\text{learning exposure t in $1,2,\dots,T$} }{
    $\theta\leftarrow\theta_{t-1}$\\
    $\mathcal{C}\leftarrow$ SELECT\_RANDOM$(\mathcal{C}\cup\mathcal{D}^{\text{train}}_t)$\\
    $\theta_t, \text{acc}_t\leftarrow$ SDFNet$(\theta,\mathcal{C},\mathcal{D}^{\text{val}}_t$)\\
 } 
\KwResult{($\text{acc}_1,\text{acc}_2,\dots\text{acc}_T$)}
\label{algo:gsmart}
\end{algorithm}

\begin{algorithm}[t!]
\caption{C-SDFNet}
\SetAlgoLined
\KwIn{Batch training procedure SDFNet$(\theta,\mathcal{D}^\text{train},\mathcal{D}^\text{val})$ that returns the trained parameters $\theta$ and the performance of the trained model on $\mathcal{D}^\text{val}$}
\KwData{(RGB image, 3D coordinates, SDF values) pair datasets $\mathcal{D}^{\text{train}}=\cup_{i=1}^T\mathcal{D}^{\text{train}}_i$,  $\mathcal{D}^{\text{val}}=\cup_{i=1}^T\mathcal{D}^{\text{val}}_i$}
\KwDef{$\ell:$ weighted $L_1$ loss }
\ForEach{\text{learning exposure t in $1,2,\dots,T$} }{
    $\theta\leftarrow\theta_{t-1}$\\
    $\theta_t, \text{acc}_t\leftarrow$ SDFNet$(\theta,\mathcal{D}^\text{train}_t,\mathcal{D}^{\text{val}}_t)$\\
 } 
\KwResult{($\text{acc}_1,\text{acc}_2,\dots\text{acc}_T$)}
\label{algo:csdfnet}
\end{algorithm}

\noindent\textbf{Loss function}: SDFNet and ConvSDFNet use $L_1$ loss as the loss function, with high weights for points close to the surface. Specifically,
\begin{equation*}
 \mathcal{L}(s,\hat{s})= 
    \begin{cases}
        \vert s-\hat{s}\vert, \quad\text{ if $\vert s \vert > 0.01$}\\
        4\vert s -\hat{s}\vert, \quad\text{ otherwise}
    \end{cases}
\end{equation*}
where $s$ is the ground truth SDF value and $\hat{s}$ is the predicted SDF value.

OccNet uses Binary Cross Entropy (BCE) loss on each input 3D point. Specifically,
\[\mathcal{L}(p,\hat{p})=-p\log\hat{p}-(1-p)\log(1-\hat{p})\] where $p\in\{0,1\}$ is the ground truth binary value and $\hat{p}$ is the predicted probability of whether a point is inside or outside the mesh.

Pointcloud Auto-encoder optimizes Chamfer distance (CD) loss on the ground truth and predicted pointclouds. Mathematically, CD loss is written as  
\[\ell(S,\hat{S})=\frac{1}{\vert S \vert}\sum_{x\in S}\min_{y\in \hat{S}}\Vert x-y\Vert_2+\frac{1}{\vert \hat{S}\vert}\sum_{y\in \hat{S}}\min_{x\in S}\Vert x-y\Vert_2\] where $S$ and $\hat{S}$ are ground truth and predicted pointclouds respectively.

\noindent\textbf{Mesh generation}: We use MISE, an algorithm that hierarchically extracts the mesh isosurface introduced by~\cite{mescheder2019occupancy} to generate the predicted mesh. Instead of generating the SDF/occupancy values for all the points uniformly sampled in the cube, MISE starts from a lower resolution and hierarchically determines the voxels that contain the mesh to subdivide until the desired resolution is reached. We adapt MISE to work on both SDF and occupancy values.

\noindent\textbf{Metric}: Following~\cite{tatarchenko2019single,thai20203d}, we use F-Score at 1\% as our main evaluation metric. We first sample 300K and 100K points respectively on the surface of the predicted mesh ($S_1$) and ground truth mesh ($S_2$). The metric is computed as the following
\[
FS@1=\frac{2\cdot prec@1\cdot rec@1}{prec@1+rec@1}
\]
where $prec@1$ is the precision at 1\%, which measures the portion of points from $S_1$ that lie within a threshold 0.01 to the points from $S_2$ (in the case where the mesh is normalized to fit in a unit cube) and $rec@1$ is the recall at 1\%, which measures the portion of points from $S_2$ that lie within a threshold 0.01 to the points from $S_1$.

\begin{figure*}[t!]
\begin{minipage}[t!]{\linewidth}
    \includegraphics[width=\linewidth]{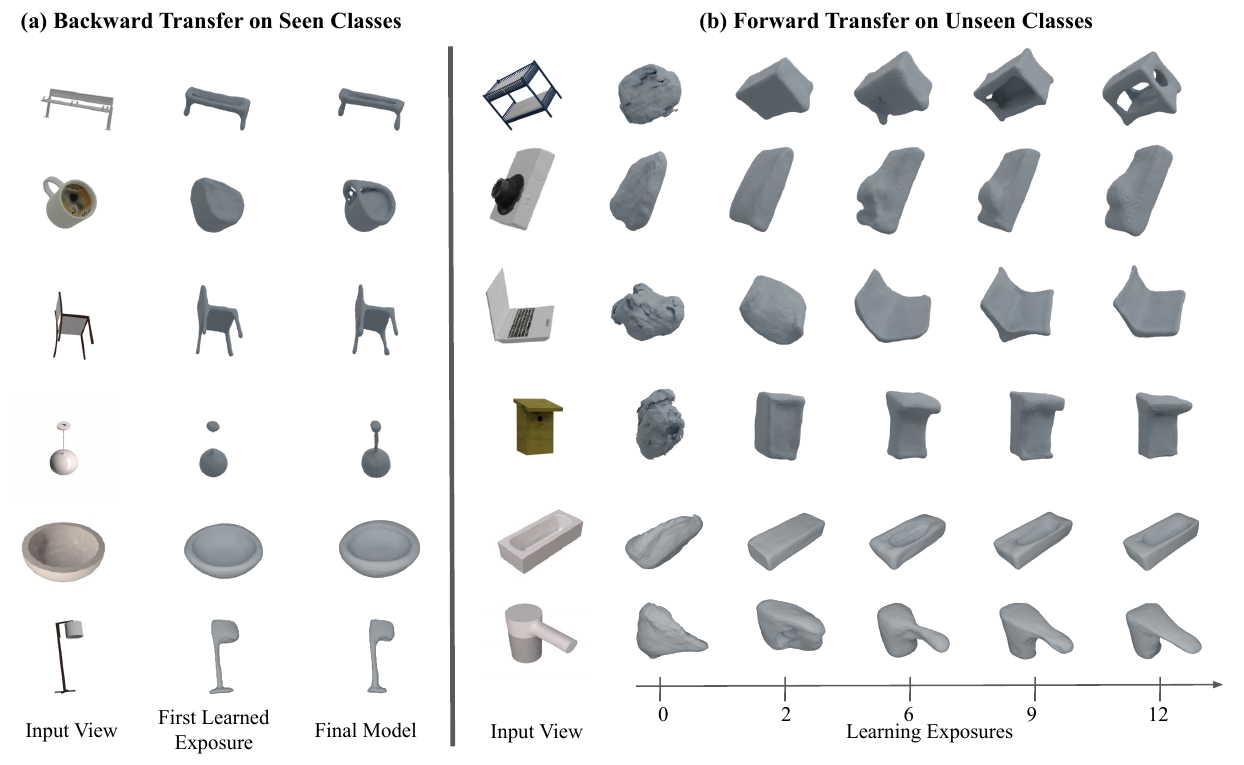}
    \caption{Additional qualitative results for CL 3D shape reconstruction: (a) Positive backward transfer, reconstructions based on the first learning exposure improve by the time the final model is trained, demonstrating lack of catastrophic forgetting, and (b) Positive forward transfer, reconstruction performance on unseen object classes improves steadily during training, demonstrating the generalization ability of shared representation. These results are obtained by vanilla SGD \emph{without} special architectures, losses, or exemplars. }
    \label{fig:bwt_fwt_sup}
\end{minipage}%

\end{figure*}

\noindent\textbf{Additional Results:} We present qualitative results in Fig.~\ref{fig:bwt_fwt_sup}. We can see that continual object 3D shape reconstruction experiences positive knowledge transfer, with improved performance on both seen and novel classes over time.

\begin{figure}[t!]
\begin{minipage}[t!]{\linewidth}
    \includegraphics[width=\linewidth]{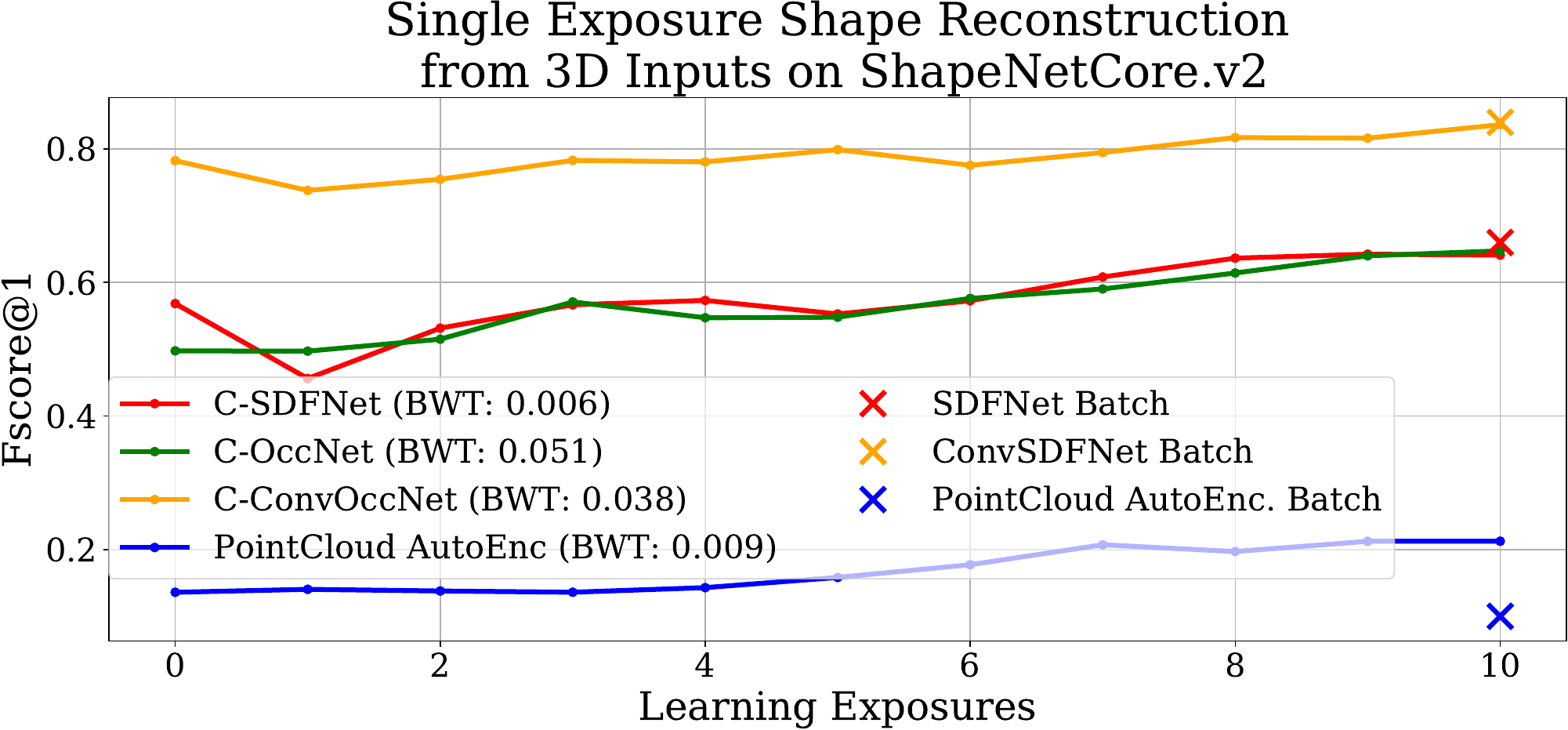}
    \caption{Performance of various 3D object shape reconstruction algorithms with 3D input on ShapeNetCore.v2, single exposure setting with 5 classes/learning exposure. All algorithms achieve positive BWT.}
    \label{fig:3dinp}
\end{minipage}%

\end{figure}

In Fig.~\ref{fig:3dinp} we present the performance of SDFNet, OccNet, ConvOccNet and Pointcloud Auto-encoder with 3D input in single exposure setting with 5 classes/exposure on 55 classes of ShapNetCore.v2. Note that all algorithms achieve positive BWT, illustrating that our findings hold on various model architectures and input/output representations.

\begin{figure}[h]
\centering
\vspace{-10pt}
\includegraphics[width=\linewidth]{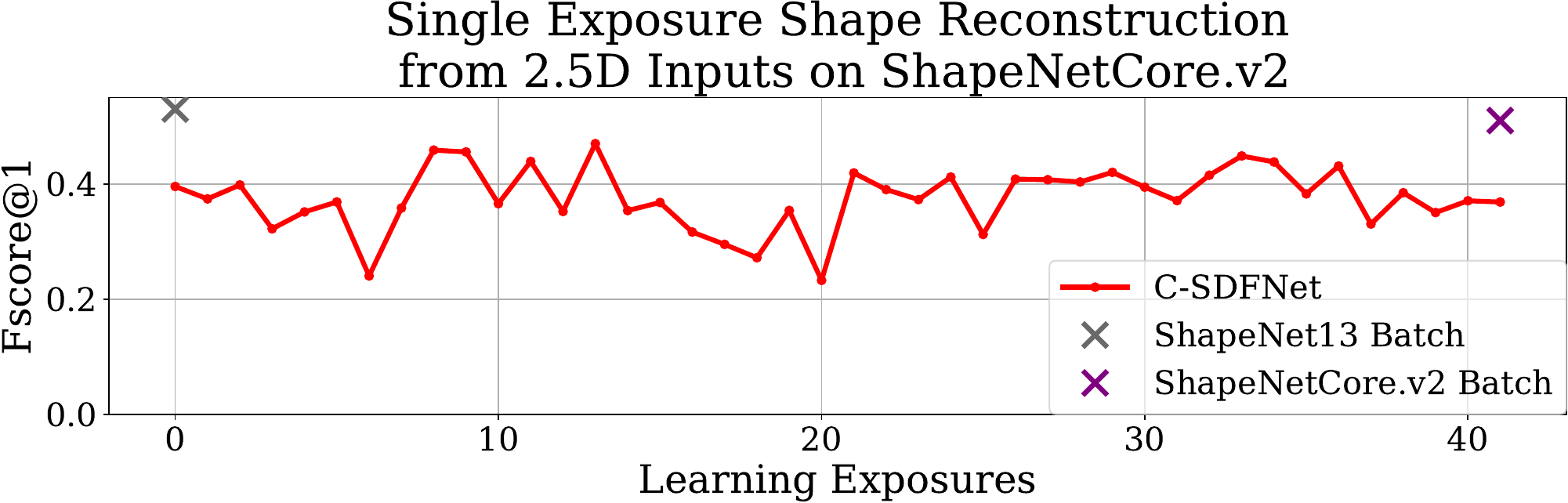}

\caption{CL performance of C-SDFNet on 42 classes, 1 class/learning exposure of ShapeNet after being trained until convergence in batch mode on ShapeNet13.
}\label{fig:cl42}
\vspace{-10pt}
\end{figure}

We further conduct an experiment where the model is trained in batch mode until convergence on 13 classes of ShapeNet13, and then continually trained on the remaining 42 classes of ShapeNet with 1 class sample/exposure. We report FScore@1 on all learned classes (including the initial 13 classes) and show the result in Fig.~\ref{fig:cl42}. The performance remains relatively constant over 42 learning exposures. We note that this is not the case in classification~\cite{stojanov2019incremental} where CL models initialized with ImageNet pretrained weights still experienced catastrophic forgetting upon continually observing new data.

\subsection{Single-view 2.5D Sketches Prediction (Sec.~4)}
\noindent\textbf{Architecture}: We adapt the 2.5D sketch estimation from MarrNet~\cite{wu2017marrnet} for continual training. The backbone architecture for MarrNet is a U-ResNet18 with the ResNet18 image encoder initialized with ILSVRC-2014 pre-trained weights.

\noindent\textbf{Loss functions}: We use MSE as the loss function for depth and normals prediction. Specifically,
\begin{align*}
    &MSE(I,\hat{I})=\frac{1}{K\times K}\sum_{i,j}^K\Vert I(i,j)-\hat{I}(i,j)\Vert_2^2
\end{align*}
where $I$ and $\hat{I}$ are the ground truth and predicted images respectively.

\noindent\textbf{Metrics}: For depth prediction, we report threshold accuracy: percentage of $y_i$ such that \[\max\left(\frac{y_i}{y_i^\star},\frac{y_i^\star}{y_i}\right)<\sigma\] where $y_i$ and $y_i\star$ are the predicted and ground truth depth values at pixel $i$ and $\sigma$ is the threshold. In our evaluation, we use $\sigma=1.25$ as in~\cite{koch2018evaluation}.

For normals, we report cosine distance threshold as the main metric. We first convert the RGB values of the normal map into 3D vectors  
\[\boldv{n}=2\left(\boldv{c}-\begin{bmatrix}0.5\\0.5\\0.5\end{bmatrix}\right)\] where $\boldv{n}$ and $\boldv{c}=\begin{bmatrix}\text{r}\\ \text{g}\\ \text{b}\end{bmatrix}$ are the normal and color vectors respectively. Cosine distance threshold accuracy is then computed as 
\[\left\langle\frac{\boldv{n}}{\Vert \boldv{n}\Vert_2},\frac{\boldv{n}^\star}{\Vert \boldv{n}^\star_2\Vert}\right\rangle >\sigma\] where $\boldv{n}$ and $\boldv{n}^\star$ are predicted and ground truth normals. We set $\sigma=0.9$.

\begin{figure*}[t!]
\begin{minipage}[t!]{\linewidth}
    \includegraphics[width=\linewidth]{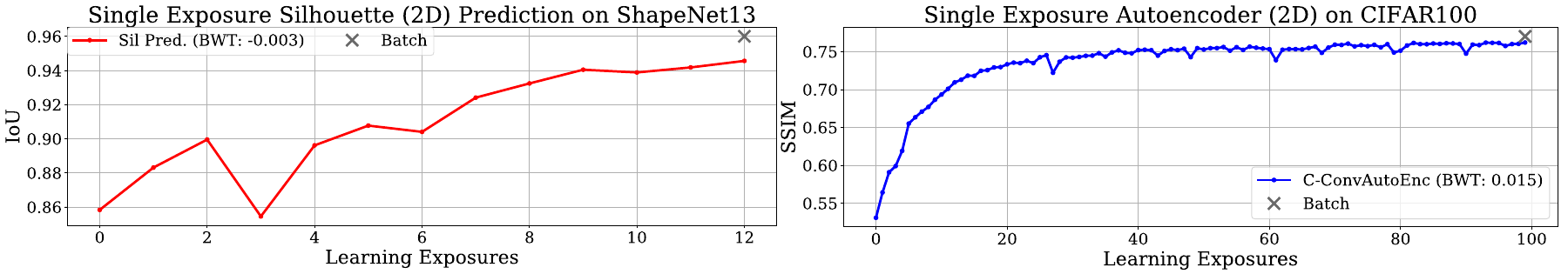}
    \caption{(left) IoU of silhouette prediction model (right) SSIM of image autoencoding. Backward transfer
is reported in parenthesis. The performance of CL models increases over time and approaches batch in 2D reconstruction task.}
    \label{fig:single_sil_dn}
\end{minipage}%

\end{figure*}

\subsection{2D Reconstruction}
We conduct additional experiments on continual 2D to 2D mapping that includes learning to segment foreground/background given an RGB input image and image autoencoding. We present results in Fig.~\ref{fig:single_sil_dn} which demonstrate that these tasks do not suffer from catastrophic forgetting.


\subsubsection{Silhouette Prediction}
We utilize U-ResNet18-based  MarrNet~\cite{wu2017marrnet} architecture train with BCE loss. We report Intersection-over-Union for silhouette prediction as the metric. Specifically,
\begin{align*}
    &IoU(I,\hat{I})=\frac{\vert I \cap \hat{I}\vert}{\vert I \cup \hat{I}\vert}
\end{align*}
The average IoU at each learning exposure (Fig.~\ref{fig:single_sil_dn} left) demonstrates that single exposure exposure silhouette prediction does not suffer catastrophic forgetting (minimal negative backward transfer). In fact we observe that the IoU increases over time.

\subsubsection{Image Autoencoding}
\noindent\textbf{Architecture}: We implement a shallow network with 4 conv. layers, each followed by a max pooling layer which we termed ConvAutoEncoder. Each conv. layer has 16 channels and the dimension of the bottle-neck feature vector is $16\times2\times2$. The network is randomly initialized.

\noindent\textbf{Loss function}: We train ConvAutoEncoder with MSE loss for each pixel, defined as
\[
\mathcal{L}(I, \hat{I})=\frac{1}{K\times K\times 3}\sum_{c=1}^3\sum_{i,j}^K\Vert I(i,j,c)-\hat{I}(i,j,c) \Vert_2^2
\]
where $K$ is the size of the input image and $c=\{1,2,3\}$ is the 3 input channels (red, green, blue).

\noindent\textbf{Metric}: We use SSIM scaled to range $[0,1]$ as the main evaluation metric for the image autoencoding experiment. Specifically, given two image windows $x$ and $y$ of the same size $N\times N$ the original SSIM metric is computed as
\[SSIM(x,y)=\frac{(2\mu_x\mu_y+c_1)(2\sigma_{xy}+c_2)}{(\mu_x^2+\mu_y^2+c_1)(\sigma_x^2+\sigma_y^2+c_2)}\]
with $\mu_x,\mu_y$ be the averages of $x$ and $y$ respectively, $\sigma^2_x,\sigma^2_y,\sigma_{xy}$ are the variances of $x,y$ and the covariance of $x$ and $y$ respectively, $c_1,c_2$ are constants to avoid dividing by $0$ in the denominator.

We experiment on CIFAR-100~\cite{krizhevsky2014cifar} (size $32\times 32$) with one class per exposure and report the average SSIM~\cite{wang2004image} as the accuracy metric at each learning exposure (Fig.~\ref{fig:single_sil_dn} right). SSIM increases over time and eventually reaches batch performance. This is yet more evidence for the robustness of continual reconstruction.

\subsection{Classification Baselines (Sec.~7)}

\noindent\textbf{GDumb}~\cite{prabhugdumb} is an algorithm that randomly selects exemplars and performs training on the exemplar set only. At each learning exposure, the model is trained from scratch on the exemplar set, in which each category is represented with the same number of samples. GDumb utilizes the standard cross-entropy loss and classifies using the network outputs. We used our PyTorch implementation of GDumb with ResNet18 initialized randomly as the feature extractor.

\noindent\textbf{Classifier with Exemplars} is a simple baseline where we train a standard classifier with cross-entropy loss continually. At each learning exposure, the learning model is trained on the current training data combined with the randomly selected exemplar set without any further heuristics. Similar to GDumb, we use randomly initialized ResNet18 as the feature extractor.  

\noindent\textbf{ImageNet Pretrained} is the baseline we use to highlight that the feature space learned by CL single-view 3D shape model from RGB image without ground truth label is discriminative. For each new class, we randomly select the exemplar set from the training data. At test time, we first extract the feature representation from the ILSVRC-2014 pretrained ResNet18 for each test sample. We then perform NCM to predict the label using the exemplar set.

\section{Effect Of Exemplar Set Size On Output Distribution Shift}\label{sec:exemplar_output}
\begin{figure}[t]
\begin{minipage}[t!]{\linewidth}
    \includegraphics[width=\linewidth]{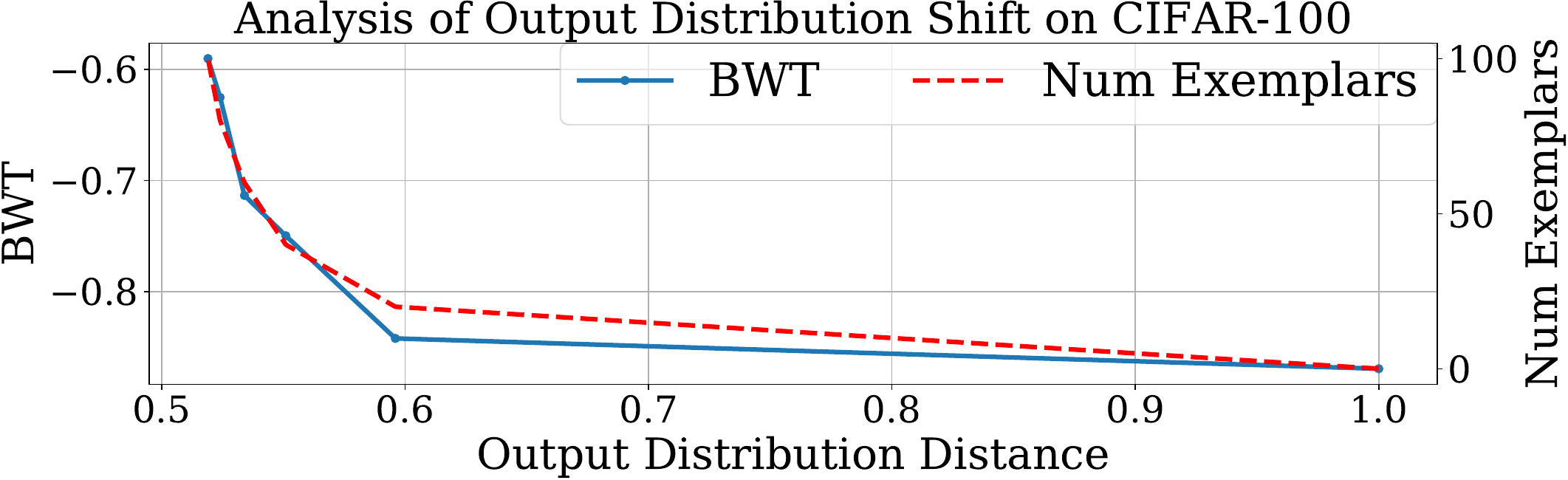}
    \vspace{-20pt}
    \caption{The relationship between the output distribution shift across learning exposures, the number of exemplars per class and BWT on CIFAR-100. Experiments are run for 0, 20, 40, 60, 80, 100 exemplars/class. Larger exemplar set size leads to smaller output distribution distance and higher BWT.}
    \label{fig:dist}
\end{minipage}%
\end{figure}

We apply the analysis technique described in Sec.~6 of the main text to gain insight into the effectiveness of replay methods commonly used to avoid catastrophic forgetting in classification. We design our experiment on CIFAR-100 with 1 class per learning exposure. We employ randomly initialized ResNet34 and vary the exemplar set size from 0 to 100 exemplars/class. Fig.~\ref{fig:dist} illustrates that larger exemplar set size associates with smaller conditional output distribution shift which results in improvement in BWT. 

\section{Further Explanation for Repeated Exposures Setting}\label{sec:rep_exp_expl}
In the repeated exposure setting, each class occurs a fixed number of times (e.g. 10 repetitions) in random order. For example, in the case of 50 classes repeated 10 times, we would first generate 500 learning exposures, and then perform a random permutation to obtain the order seen by the learner. As a result, classes repeat in complex and highly-variable patterns. Note that even though classes repeat, each learning exposure still contains only a single class (or a small number), thereby preserving the domain shift between exposures that makes CL challenging. 

\begin{figure*}[t!]
\centering
\begin{subfigure}{0.49\linewidth}
    \includegraphics[width = \linewidth]{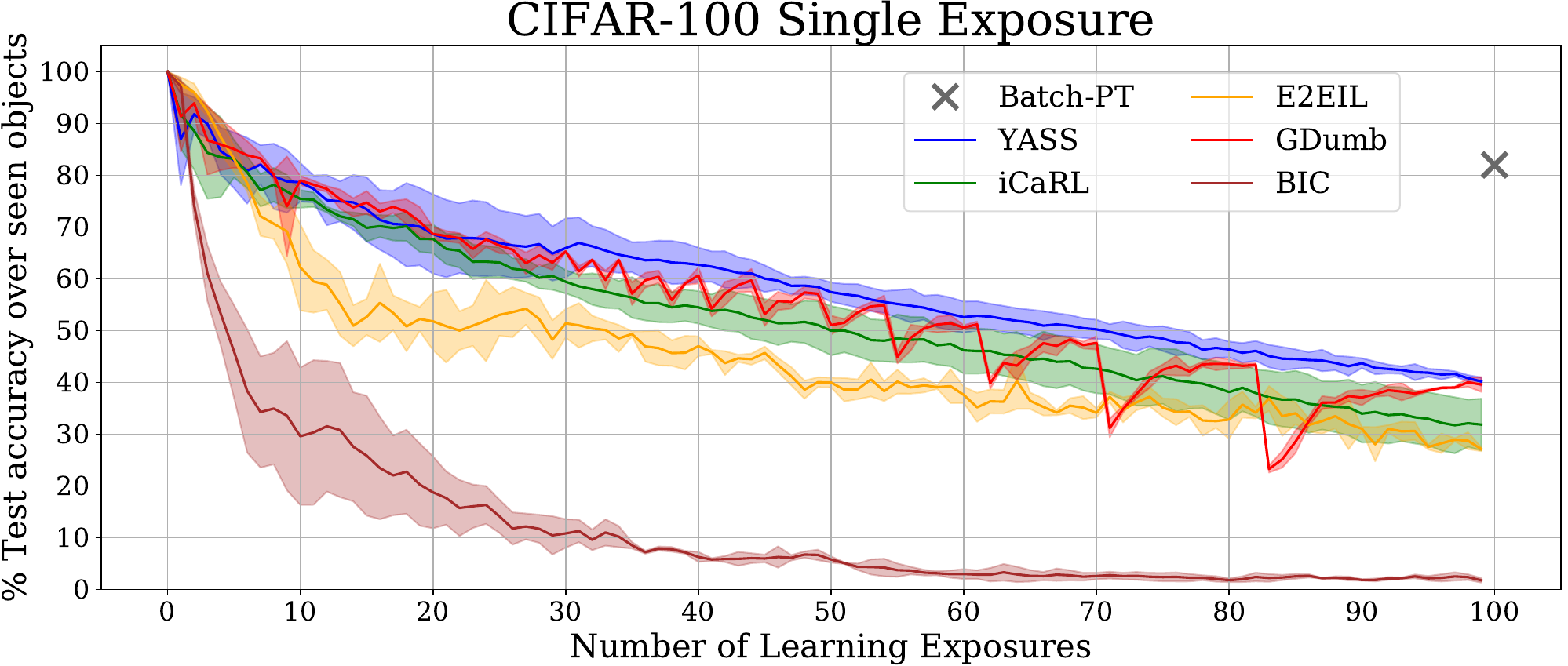}
    \caption{\label{fig:exp_cifar_single}}
\end{subfigure}
\hfill
\begin{subfigure}{0.49\linewidth}
    \includegraphics[width=\linewidth]{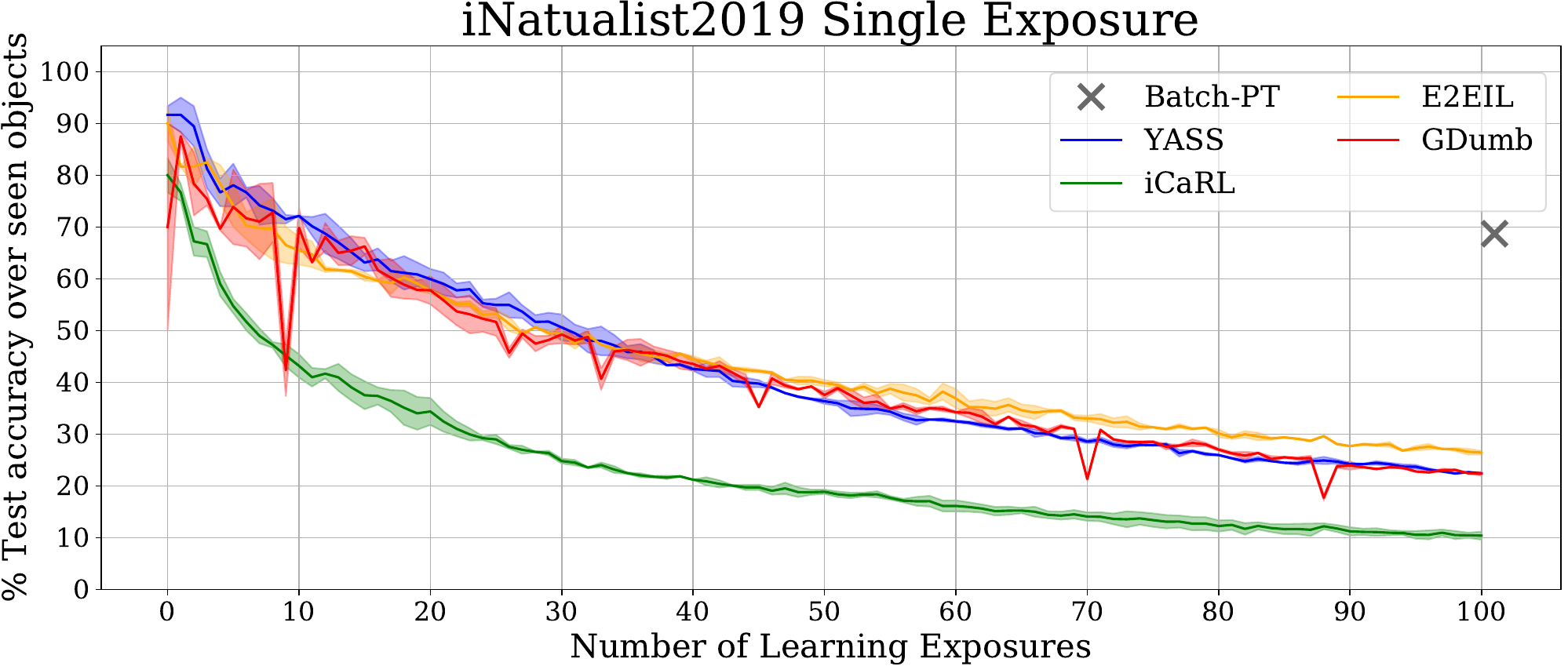} 
    \caption{\label{fig:exp_inat_single}}
\end{subfigure}
\vspace{-0.7em}
\caption{(\subref{fig:exp_cifar_single}) Performance of YASS, iCaRL~\cite{rebuffi2016icarl}, E2EIL~\cite{castro2018end}, GDumb~\cite{prabhugdumb}, and BiC~\cite{wu2019large} when presented with a single exposure for each category from CIFAR-100 with 1 class learned per exposure. Performance is averaged over 3 runs with random class orderings.
(\subref{fig:exp_inat_single}) ~YASS, iCaRL, E2EIL and GDumb on iNaturalist2019 in a single exposure setting with 10 classes learned per exposure. Performance is averaged over 2 runs. YASS outperforms others on CIFAR-100 and achieves competitive performance on iNaturalist-2019.
}\label{fig:cifar_classification}
\vspace{-17pt}
\end{figure*}

\begin{figure}[t!]
\begin{minipage}[t!]{\linewidth}
    \includegraphics[width=\linewidth]{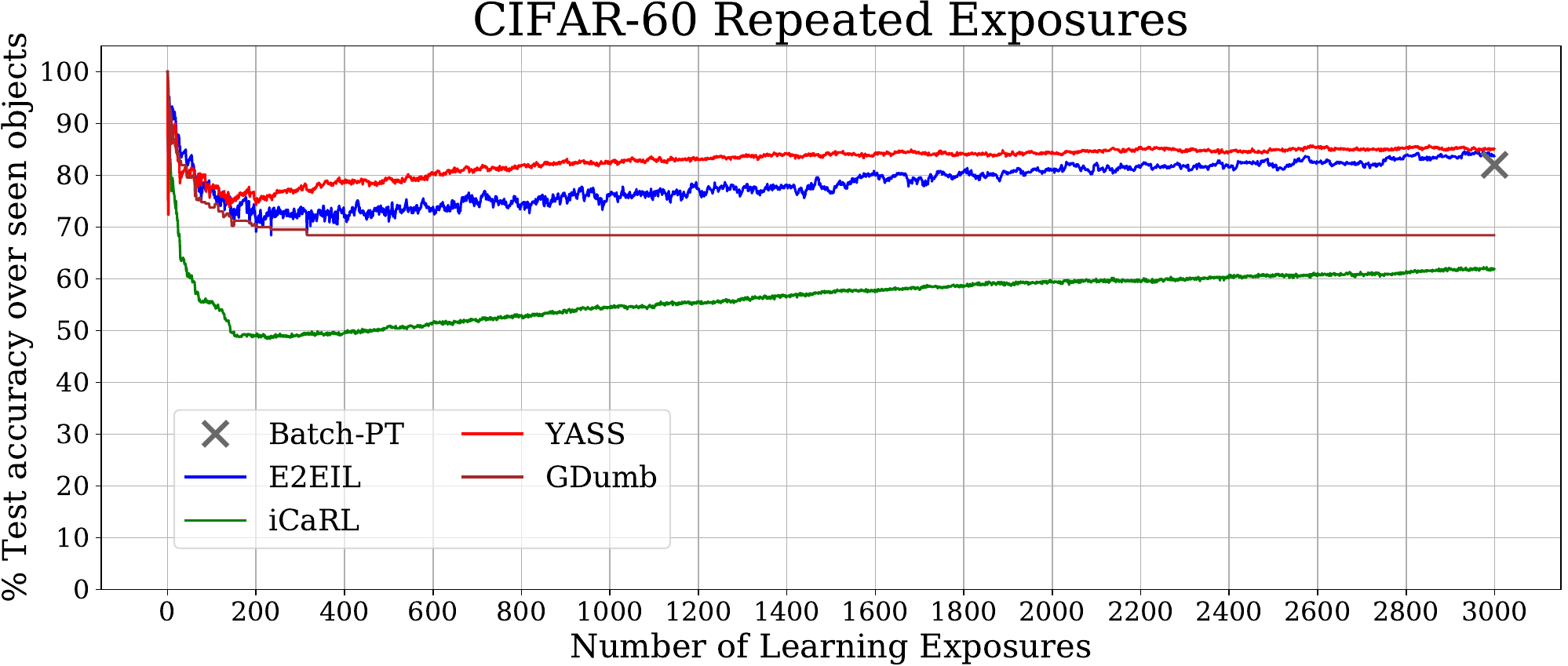}
    \caption{Performance of YASS, iCaRL, E2EIL and GDumb on CIFAR-60 with repeated exposures. YASS demonstrates a strong performance over SOTA methods.}
    \label{fig:cifar_repeated}
\end{minipage}%

\end{figure}

\section{YASS---Simple Baseline for Classification}\label{sec:yass}
Our findings have highlighted the robustness to forgetting in continual reconstruction, making it clear that effective approaches to incremental representation learning in classification remain a key challenge. In this section, we address the question of what exactly are the key ingredients in an effective approach to continual classification? Inspired by the work of~\cite{prabhugdumb}, we present a simple baseline method for \emph{class-incremental} classification, which we term \emph{YASS} (Yet Another Simple baSeline). YASS encapsulates a minimal set of algorithmic components: using exemplars chosen at random, employing weighted gradient for balancing, classifying via the network output with cross-entropy loss, and importantly, applying continuous representation learning approach. YASS adapts standard batch learning to the continual learning context with the fewest changes. Surprisingly, we show that YASS achieves competitive performance in the class-incremental single task setting.

\noindent\textbf{Exemplar Memory and Management}. As in \cite{castro2018end, rebuffi2016icarl,prabhugdumb,wu2019large} we allow for a small (less than $3\%$ of total training data) exemplar set. Similar to \cite{prabhugdumb}, rather than using complex heuristics like herding~\cite{rebuffi2016icarl}, we use random exemplar selection where the exemplar samples are randomly chosen from the learning data. In prior memory based algorithms the exemplar set size is fixed and equally split among all learned concepts which leads to unused memory. For example, when the exemplar set size is 2000 images, after the 91\textsuperscript{st} concept is learned, each concept will evenly have 21 exemplars, which leaves 89 exemplar slots in the memory unused. To counter this issue, we equally divide the remaining exemplar slots to the first learned concepts.

\noindent\textbf{With data balancing}. To address the issue where the new training data significantly outnumbers stored exemplar data, we propose a data balancing mechanism based on a common method as described in \cite{japkowicz2002class} and refer to it as Weighted Gradient (WG). Specifically, we make sure that every class in the training data contributes equally during backpropagation by scaling the gradients inversely proportionally to the number of samples of the corresponding class.

\begin{figure*}[t!]
\centering
\begin{subfigure}{0.49\linewidth}
    \includegraphics[width = \columnwidth]{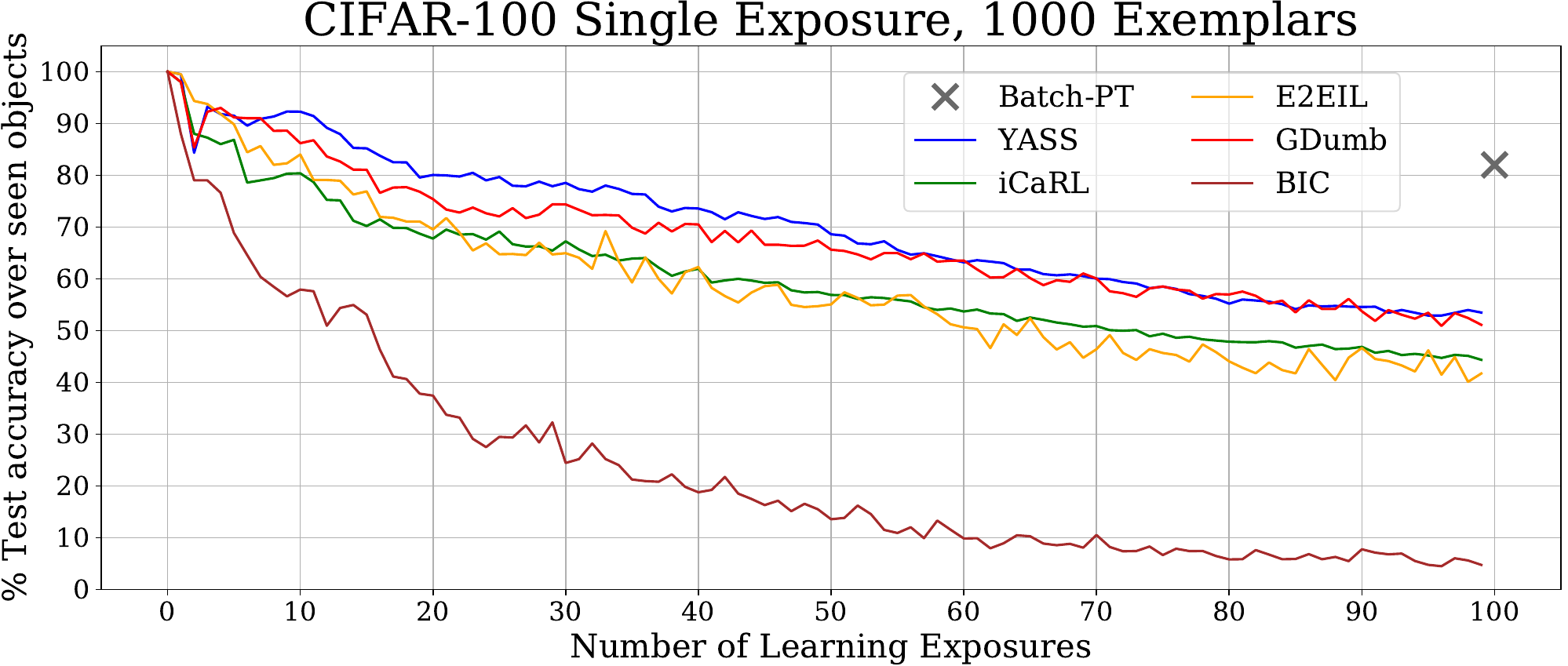}
    \caption{\label{fig:cifar_1000}}
\end{subfigure}
\begin{subfigure}{0.49\linewidth}
    \includegraphics[width=\linewidth]{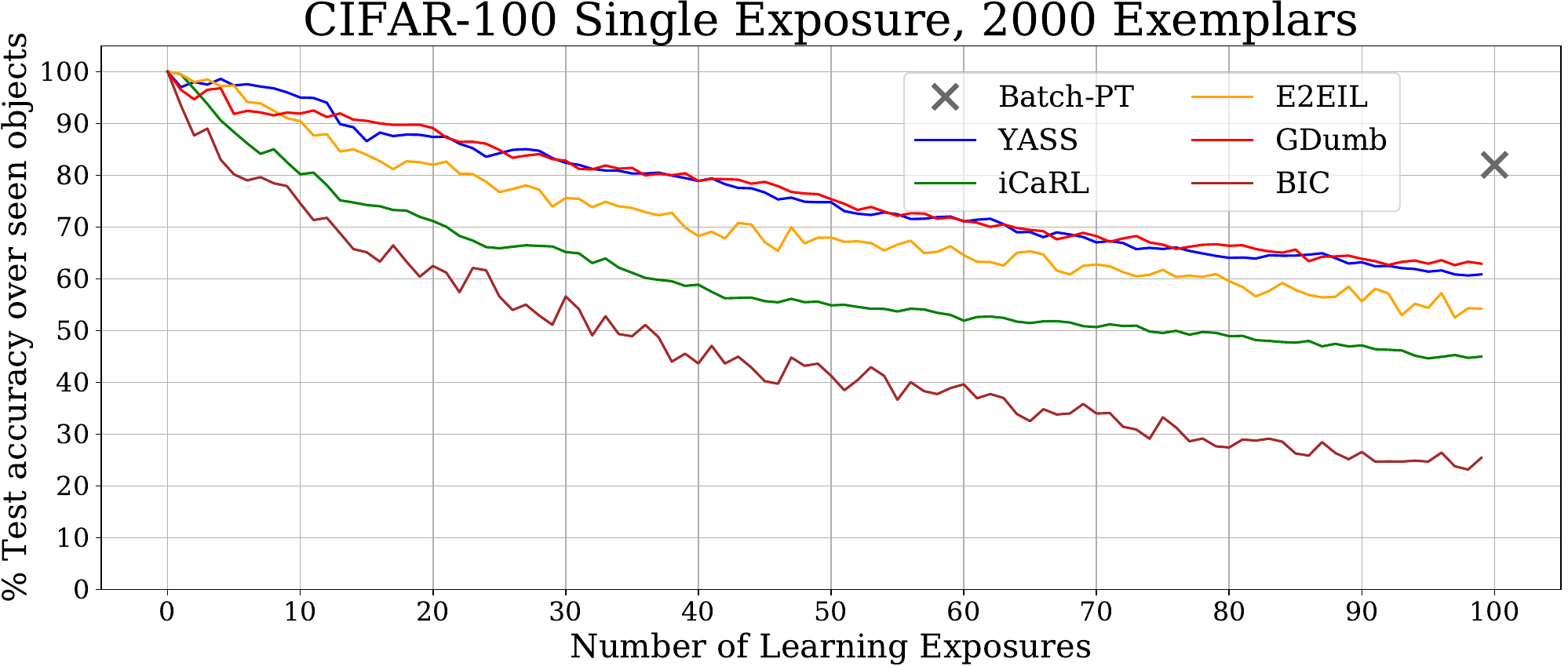} 
    \caption{\label{fig:cifar_2000}}
\end{subfigure}
\vspace{-0.7em}
\caption{(\subref{fig:cifar_1000}) Performance of YASS, iCaRL, E2EIL and GDumb on CIFAR-100 in the single exposure case with 1000 exemplars. 
(\subref{fig:cifar_2000}) ~with 2000 exemplars. YASS demonstrates a consistently strong performance over SOTA methods.
}\label{fig:cifar_explr}
\end{figure*}

\noindent\textbf{Experiments}. We conduct experiments on two image datasets: CIFAR-100~\cite{krizhevsky2009learning} and the challenging large scale dataset, iNaturalist-2019~\cite{vanhorn2018inaturalist} with 1010 categories of highly similar species. We evaluate the performance of YASS against competitive baselines for class-incremental subtask that allow exemplar memory, classified as formulation B2 in~\cite{prabhugdumb}. Despite the simple design choices, YASS outperforms these methods on the most challenging setting of CIFAR-100 (one class learned per exposure with 500 exemplars) in Fig.~\ref{fig:exp_cifar_single} and achieves competitive performance on iNaturalist-2019 dataset with 10 classes learned per exposure and 8080 exemplars in Fig.~\ref{fig:exp_inat_single}. 

We provide additional evidence that YASS outperforms other baselines on CIFAR-60 dataset with repeated exposures. In this experiment, each of the 60 classes is present 50 times. The exemplar set size is 1600, which is approximately 5.3\% of the training set. We compare YASS against iCaRL~\cite{rebuffi2016icarl} and E2EIL~\cite{castro2018end} as in~\cite{stojanov2019incremental}. YASS outperforms these methods in the repeated exposures case (Fig.~\ref{fig:cifar_repeated}). Since YASS, E2EIL and iCaRL are continuous representation learning approaches (discussed in Sec.~\ref{sec:fwt_sup}), the feature representation is refined when a category is exposed again and thus, demonstrating an increasing trend in the performance and eventually reaching that of the batch model (for YASS and E2EIL). Additionally, we compare YASS against GDumb, an episodic representation learning approach. Since GDumb is trained from scratch at each learning exposure, the feature representation does not benefit from repetition. YASS and E2EIL outperform GDumb by $15\%$ at the end, demonstrating the advantage of the continuous over the episodic representation learning approach in the repeated exposures case.

We further demonstrate the consistently strong performance of YASS with different number of exemplar set sizes (Fig.~\ref{fig:cifar_explr}). We evaluate the performance of different methods on CIFAR-100 in the single exposure case, with 1000 exemplars (Fig.~\ref{fig:cifar_1000}) and 2000 exemplars (Fig.~\ref{fig:cifar_2000}). YASS outperforms iCaRL, E2EIL and BiC in both cases. GDumb shows a significant benefit from having more exemplars, as its performance approaches that of YASS when we increase the number of exemplars allowed.

\noindent\textbf{Continuous Representation Discussion}. YASS employs continuous representation learning, which is presumably one of the keys to success. Conversely, GDumb is an episodic representation learner, designed to test the hypothesis that there is no value in representation propagation. The lack of benefit presumably arises because the biases introduced by domain shift and catastrophic forgetting outweigh the benefits.  Sec.~\ref{sec:cl_tasks} shows that reconstruction tasks demonstrate the benefit of continuous representation learning, as they do not suffer from catastrophic forgetting. While GDumb achieves competitive performance on classification task, it is not beneficial for shape learning (Fig.~\ref{fig:fwt_sup}).

For classification, solely learning the representation continuously might not be sufficient. We train GSmart (Sec.~\ref{sec:fwt_sup}) for the classification case and find its performance to be poor. Different from prior continuous representation learning approaches like iCaRL, BiC or E2EIL, YASS allows the representation to be optimized over time instead of constraining the weights of the model on those learned from the previous exposures (eg. distillation loss), which might be a local minimum for all training data. By further carefully managing the exemplar set and making use of all the available training data with a data balancing strategy, YASS successfully illustrates the benefit of continuous feature representation learning for classification task and consistently achieves good performance in different settings. 

\begin{figure}[t!]
\begin{minipage}[t!]{\linewidth}
    \includegraphics[width=\linewidth]{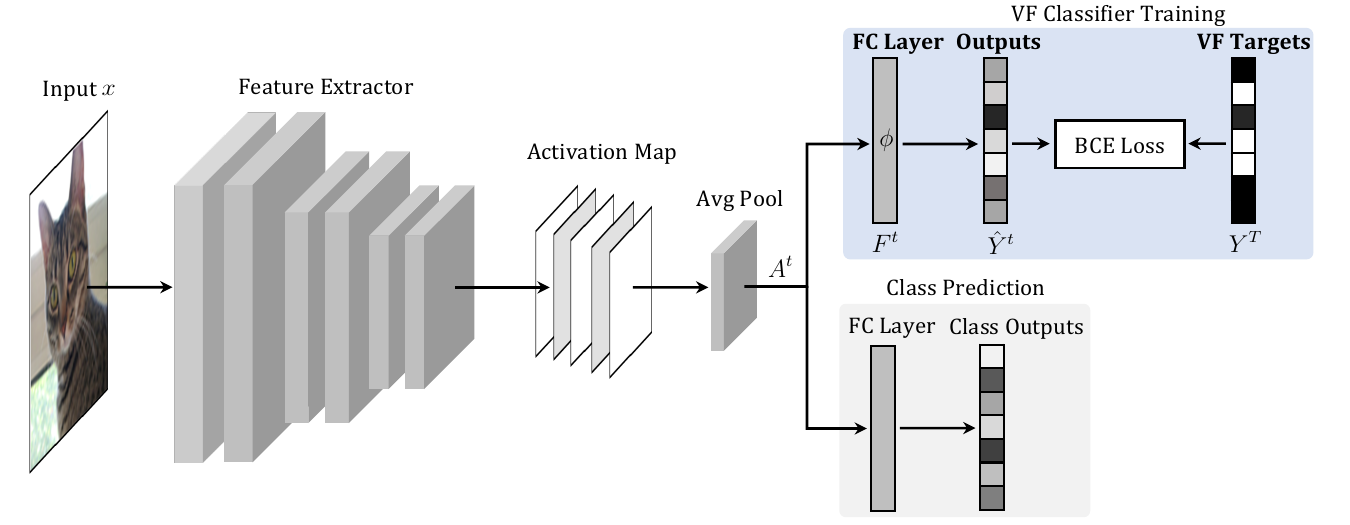}
    \caption{Architecture for the visual feature analysis. We freeze the weights of the feature extractor up to the pooling layer before the FC layer that outputs the class predictions. We train the VF classifier by optimizing the parameters $\phi_t$ of the FC layer $F^{(t)}$ (blue branch, top). The VF targets are obtained by binarizing the feature representation of the batch model. We utilize the binary cross entropy loss on each element of the predicted VF outputs $\hat{Y}^{(t)}$ and the ground truth VF targets $Y^{(B)}$. Please refer to the text for more details.}
    \label{fig:dyrt}
\end{minipage}%

\end{figure}

\begin{figure*}[t!]
\centering
\begin{subfigure}{0.49\linewidth}
    \includegraphics[width = \columnwidth]{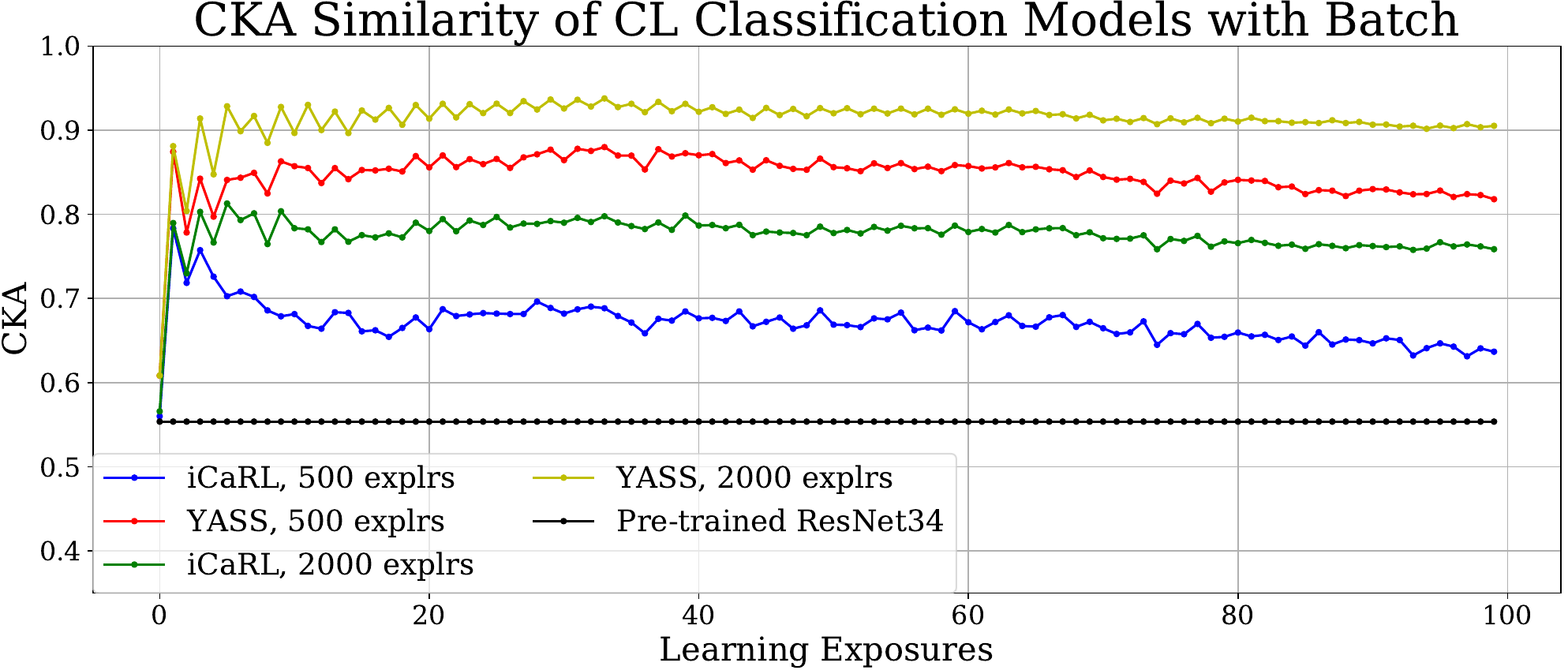}
    \caption{\label{fig:forget_a}}
\end{subfigure}
\hfill
\begin{subfigure}{0.49\linewidth}
    \includegraphics[width=\linewidth]{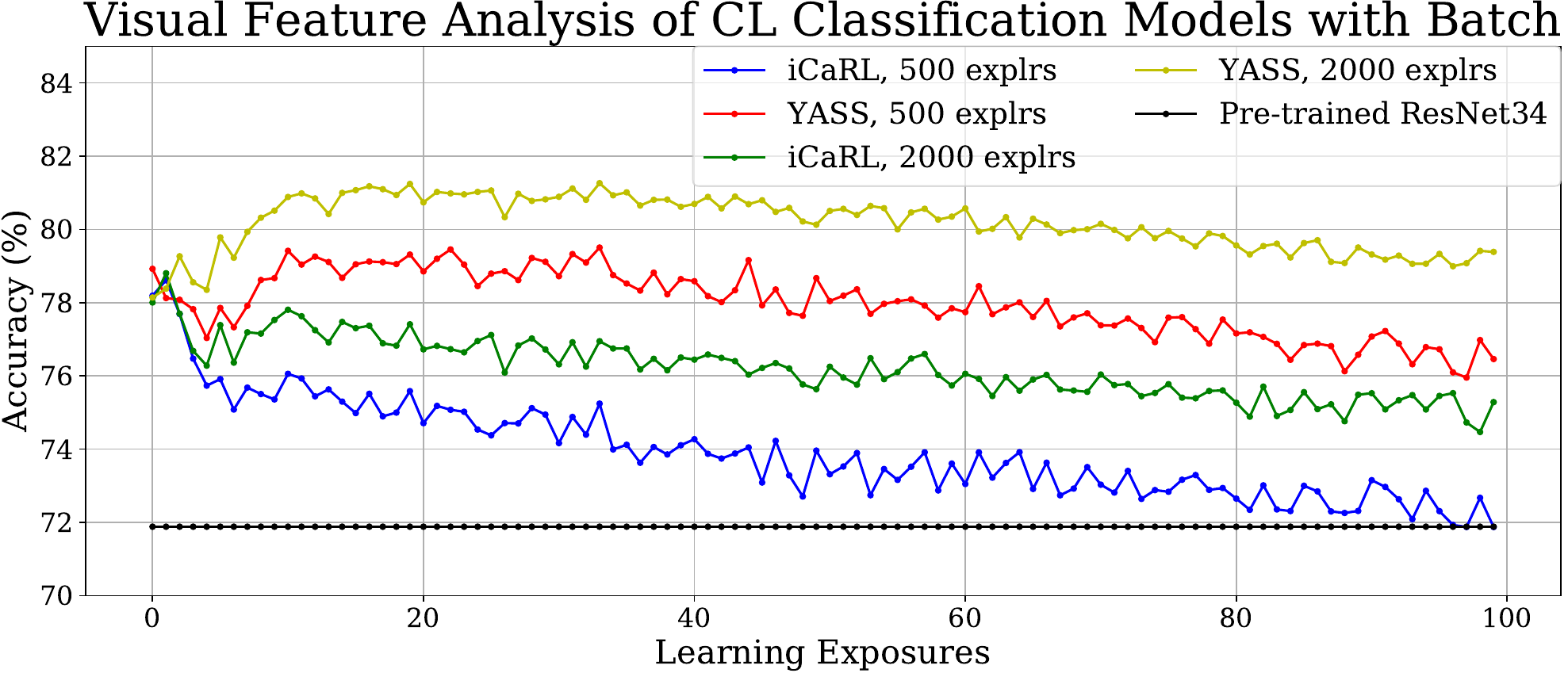} 
    \caption{\label{fig:forget_b}}
\end{subfigure}
\vspace{-0.3em}
\begin{subfigure}{0.49\linewidth}
    \includegraphics[width = \columnwidth]{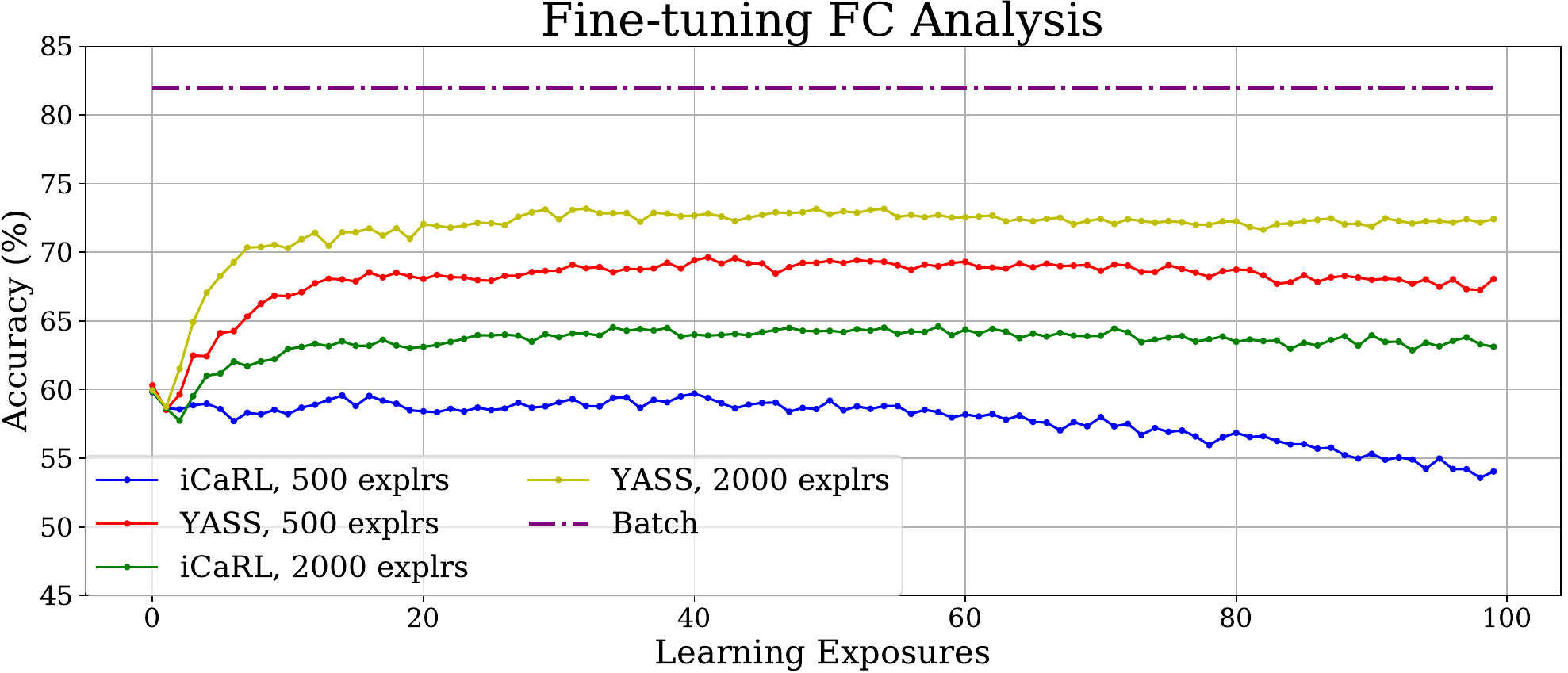}
    \caption{\label{fig:forget_c}}
\end{subfigure}
\hfill
\begin{subfigure}{0.49\linewidth}
    \includegraphics[width=\linewidth]{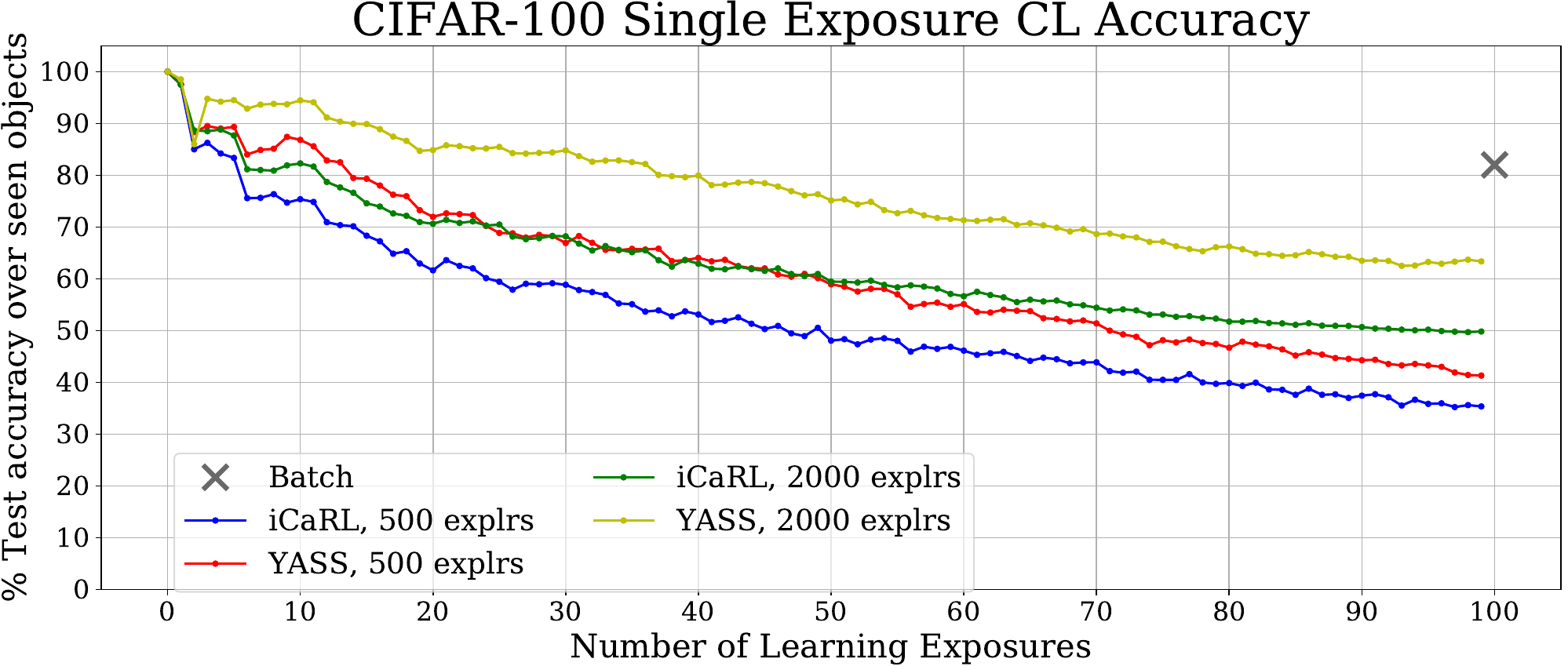} 
    \caption{\label{fig:forget_d}}
\end{subfigure}
\vspace{-0.3em}
\caption{(\subref{fig:forget_a}) CKA similarity analysis 
(\subref{fig:forget_b})~VF analysis, (\subref{fig:forget_c})~Fine-tuning FC analysis and (\subref{fig:forget_d}) CL accuracy where the average performance of learned class is plotted at each learning exposure of different CL classification and batch models. Pre-trained ResNet34 indicates the representation extracted from ILSVRC-2014 pretrained ResNet34, which is the lower bound on the performance for CKA and VF. Feature representations of YASS and iCaRL perform significantly more stable over time in all cases compared to the CL accuracy.
}\label{fig:forget}
\vspace{-15pt}
\end{figure*}

\section{Feature Representation Learning Analysis}\label{sec:dyrt}
In this section, we analyze the dynamics of forgetting in the feature representation of CL classification. While prior works demonstrated that the FC layer is susceptible to forgetting due to domain shift during CL, we believe we are the first to thoroughly investigate and provide converging evidence on the forgetting of the evolving feature representation during class-incremental classification.



We utilize the feature representation of the model learned on all training data (batch model) as the ``oracle".  We first extract the feature representation produced by the pooling layer prior to the final FC layer of both the CL models at each learning exposure and the batch model. Then, we compare these feature representations using CKA~\cite{kornblith2019similarity} and our novel visual feature analysis. We further conduct an experiment in which the feature representation at each learning exposure is frozen and the FC layer is finetuned on all training data from both learned and future classes (Fine-tuning FC).

\subsection{CKA Similarity Analysis}
Centered Kernel Alignment (CKA) introduced in~\cite{kornblith2019similarity} is a feature representation similarity measurement. Specifically, given feature representations $X$ and $Y$ with $N$ neurons, CKA is computed as
\[\text{CKA}(X,Y)=\frac{\text{HSIC}(X,Y)}{\sqrt{\text{HSIC}(X,X)~\text{HSIC}(Y,Y)}}\]
where $\text{HSIC}$ is the Hilbert-Schmidt Independence Criterion~\cite{gretton2005measuring}. CKA similarity outputs range from $0$ to $1$, where $0$ and $1$ indicate the least and the most correlation between the feature representations respectively. In our experiments, we use the RBF CKA.

\subsection{Visual Feature Analysis}
The architecture for training the visual feature analysis approach is illustrated in Fig.~\ref{fig:dyrt}. Given an input image, we first obtain the visual feature (VF) targets $Y^{(B)}$. This is done by binarizing the output of the average pooling layer $A^{(B)}$ using $\mathds{1}\{a_i^{(B)} > \theta\}$ of the batch model, where $a_i$ is each activation and $\theta$ is the threshold\footnote{Prior work on network quantization~\cite{courbariaux2016binarized} has demonstrated that activations can be quantized without incurring a significant drop in accuracy.}. For the experiments conducted in this section, we utilized threshold $\theta=1$. The VF target with value $1$ indicates that the visual feature is active and $0$ otherwise. Our goal is to train a set of $N$ binary classifiers where $N$ is the number of visual features. After obtaining the feature representation learned at each learning exposure, we then freeze the weights of the feature extractor and train the VF classifier by optimizing the parameters $\phi_t$ of the FC layer $F^{(t)}$ to produce the VF prediction $\hat{Y}^{(t)}$ (blue branch in Fig.~\ref{fig:dyrt}). Note that $F^{(t)}$ is different from the FC layer that outputs the class prediction (gray branch in Fig.~\ref{fig:dyrt}). We use binary cross entropy loss on each element of the predicted VF outputs $\hat{Y}^{(t)}$ and the ground truth VF targets $Y^{(B)}$. The intuition is that the accuracy of the VF classifiers measures the extent to which the current learned representation captures information related to the final representation.

\subsection{Finetuning FC Analysis}
Given a trained model at each learning exposure $t$, we freeze the weights of the feature extractor up to the last pooling layer before the FC layer that produces the class outputs and train a new FC layer on all training data. This includes the data from the classes learned up to exposure $t$ as well as the future classes. Note that this experiment is different from the experiments done in~\cite{wu2019large} since they only train on the data of the classes learned \emph{up to} exposure $t$. Since the FC layer is trained with the same data at each learning exposure as the batch model, the performance of Fine-tuning FC indicates the robustness of the feature representation over time compared to the batch model. 

Figs.~\ref{fig:forget_a}, \ref{fig:forget_b} show that the feature representation learned by CL models do not suffer from catastrophic forgetting as much as the class outputs from the FC layer (as compared to the significant downward trend of the curves in Fig.~\ref{fig:forget_d}). We confirm this finding by the result from Finetuning-FC (Fig.~\ref{fig:forget_c}) with the performance on all CL models very close to batch performance. Interestingly, while Figs.~\ref{fig:forget_a},~\ref{fig:forget_b},~\ref{fig:forget_c} demonstrate that the feature representation learned by YASS with 500 exemplars is more similar to the batch model than iCaRL with 2000 exemplars (red vs green curves), the CL average accuracy shows an opposite trend where YASS with 500 exemplars performs worse than iCaRL with 2000 exemplars over time.

To generate the curves in the analyses, we compare the feature representations obtained from the model trained on all training data (batch model) and the ones from the CL models at each learning exposure. Given the trained batch model, we extract the feature representation $A^{(B)}$ produced by the last pooling layer before the FC layer that outputs the class predictions. The representation learned by the CL models at each learning exposure $t$, $A^{(t)}$ is obtained in a similar way.

\end{document}